\documentclass[lettersize,journal]{IEEEtran}
\usepackage{amsmath,amsfonts}
\usepackage{algorithmic}
\usepackage{algorithm}
\usepackage{array}
\usepackage[caption=false,font=normalsize,labelfont=sf,textfont=sf]{subfig}
\usepackage{textcomp}
\usepackage{stfloats}
\usepackage{url}
\usepackage{verbatim}
\usepackage{graphicx}
\usepackage{cite}
\usepackage{epstopdf}
\usepackage[hidelinks]{hyperref}
\usepackage[capitalize]{cleveref}
\usepackage{subfig}
\usepackage{pdfpages}
\usepackage{tabularx}
\usepackage{multirow}
\usepackage{multicol}
\usepackage{url}
\usepackage[utf8]{inputenc}
\usepackage[small]{caption}
\usepackage{graphicx}
\usepackage{bbm}
\usepackage{amssymb}
\usepackage{cleveref}
\usepackage{booktabs}
\usepackage{graphicx}
\usepackage{threeparttable}

\usepackage{pifont}
\usepackage{bbding}
\usepackage{caption} 
\begin{document}
	\title{Addressing Skewed Heterogeneity via Federated Prototype Rectification with Personalization}
	
	\author{Shunxin~Guo, Hongsong~Wang, Shuxia~Lin,~Zhiqiang~Kou,~and~Xin Geng,~\IEEEmembership{Senior~Member,~IEEE}
		\thanks{This research was supported by the National Science Foundation of China (62125602, 62076063, 62302093). (\textit{Corresponding authors: Hongsong Wang; Xin Geng}).}
		\thanks{
			The authors are with the School of Computer Science and Engineering, Southeast University, Nanjing 211189, China, and also with the Key Laboratory of New Generation Artificial Intelligence Technology and Its Interdisciplinary Applications (Southeast University), Ministry of Education, China. (e-mail: sxguo@seu.edu.cn; hongsongwang@seu.edu.cn; shuxialin@seu.edu.cn; zhiqiang\_kou@seu.edu.cn; xgeng@seu.edu.cn).}
	}
	
	\markboth{Journal of \LaTeX\ Class Files,~Vol.~14, No.~8, August~2021}%
	{Shell \MakeLowercase{\textit{et al.}}: A Sample Article Using IEEEtran.cls for IEEE Journals}

	\maketitle
	
	\begin{abstract}
		Federated learning is an efficient framework designed to facilitate collaborative model training across multiple distributed devices while preserving user data privacy. A significant challenge of federated learning is data-level heterogeneity, i.e., skewed or long-tailed distribution of private data. Although various methods have been proposed to address this challenge, most of them assume that the underlying global data is uniformly distributed across all clients. This paper investigates data-level heterogeneity federated learning with a brief review and redefines a more practical and challenging setting called Skewed Heterogeneous Federated Learning (SHFL). Accordingly, we propose a novel Federated Prototype Rectification with Personalization which consists of two parts: Federated Personalization and Federated Prototype Rectification. The former aims to construct balanced decision boundaries between dominant and minority classes based on private data, while the latter exploits both inter-class discrimination and intra-class consistency to rectify empirical prototypes. Experiments on three popular benchmarks show that the proposed approach outperforms current state-of-the-art methods and achieves balanced performance in both personalization and generalization. 
	\end{abstract}
	
	\section{Introduction}
	Federated Learning (FL) facilitates collaboration among multiple clients to collectively develop a global model, effectively addressing the challenge of data silos while preserving client privacy~\cite{mcmahan2017communication,2022Tan,FedAUX2023,gu2021privacy}. It has garnered significant attention across a diverse array of industrial applications~\cite{liu2020fedvision,zeng2021multi}, including smart homes~\cite{aivodji2019iotfla,li2020review}, autonomous vehicles~\cite{zeng2021multi}, and medical diagnostics{\color{black}~\cite{dong2020can, LIU2023109739,chen2023metafed}}.
	
	\begin{figure}[!htb]
		\begin{center}
			\includegraphics[width=9cm,height=7cm]{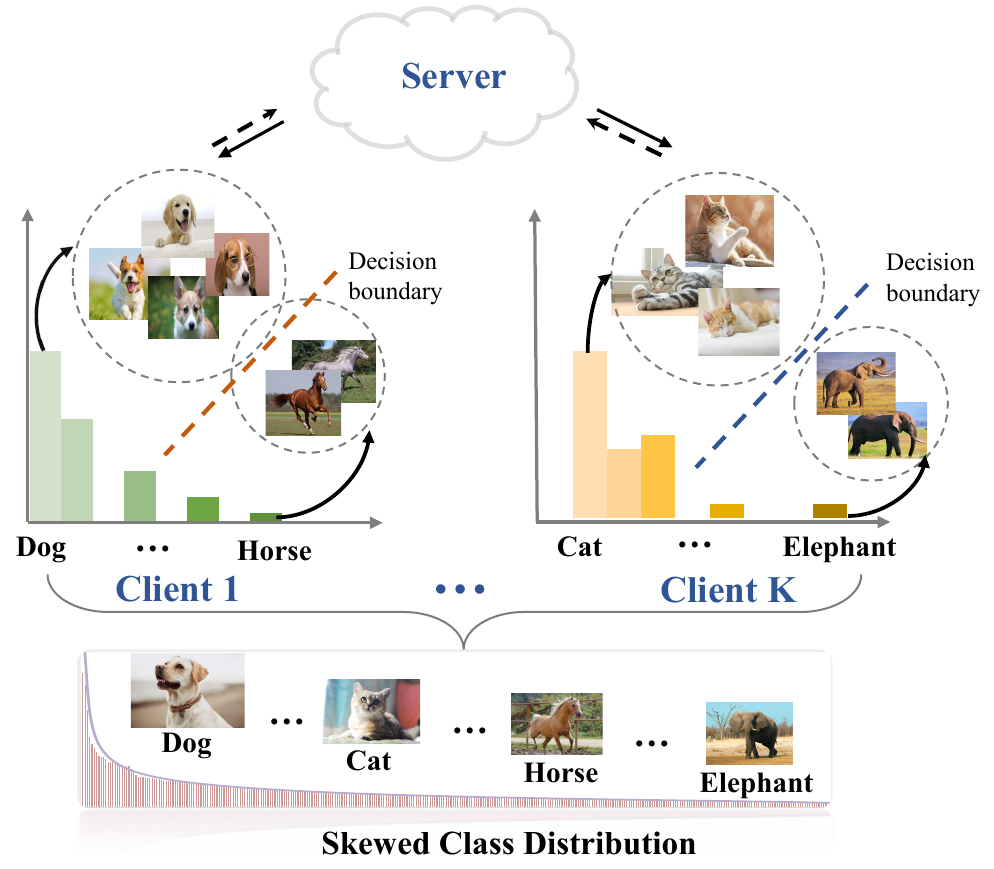}
		\end{center}
		\caption{\textbf{Illustration of skewed heterogeneous federated learning.} Each participant contains a unique skewed class distribution, while the decision boundary is biased toward the dominant classes. The heterogeneity across participants leads to inconsistencies in the decision boundaries, which further affects the aggregation update of the global model.}
		\label{fig:intro}
	\end{figure}
	
	{\color{black}As channels through which edge devices collect image data are often diverse, the non-independent and identically distributed (non-iid) data heterogeneity issue exists among different devices{\color{black}~\cite{gu2021privacy,shi2023towards,li2024dsfedcon, dong2024fadngs}}, i.e., inconsistent class distributions and differing decision boundaries.} Such data heterogeneity leads to significant discrepancies in local model parameters across devices, which results in a sharp decline in the performance of the aggregated global model.
	Many approaches have been proposed to tackle this issue, e.g., correcting the label distribution deviation from a statistical perspective~\cite{zhang2022federated,chen2020fedbe}, constructing the personalized model based on representation learning~\cite{finn2017model, li2019fedmd, wang2019federated,liang2020think, collins2021exploiting, li2021ditto}, and distilling global knowledge into local models~\cite{afonin2021towards,zhang2022fine}. Although these methods mitigate the impact of data heterogeneity and improve the generalizability of the global model, they assume that the overall data is in a balanced distribution, which is not well suited to real-world scenarios.
	
	In practical applications, the training samples commonly exhibit a skewed class distribution, where a limited number of classes are associated with a substantial volume of samples, while others are linked to only a few samples~\cite{zhang2023deep}. This phenomenon is particularly pronounced in domains such as healthcare, where the prevalence of different diseases varies significantly, and the distribution of disease samples across medical institutions is skewed.
	Several approaches have been employed to address this issue, including leveraging knowledge assistance from unlabeled or balanced proxy datasets~\cite{2022Federated,afonin2021towards,yang2023integrating}, implementing traditional class re-balancing techniques~\cite{cao2019learning, dredze2010multi, dredze2008online, yang2021federated}, or redesigning the network architecture itself~\cite{samuel2021distributional}.
	
	{\color{black}This paper systematically revisits the most practical and challenging problem in FL, i.e., Skewed Heterogeneous Federated Learning (SHFL). 
		As illustrated in Figure~\ref{fig:intro}, non-iid data leads to inconsistent class distributions across clients~\cite{kairouz2021advances,huang2022learn}. 
		Simultaneously, the skewed distribution of classes results in a decision boundary that is biased towards the dominant class. Since the prevalent approach for deep learning classification involves utilizing a linear classifier \(p = \varphi(w^\top f + b)\), where \(\varphi\) denotes the softmax function and the bias term \(b\) can be discarded, the imbalanced feature often results in the classifier weight norm of the class with more samples being greater than that of the class with fewer samples~\cite{zhang2023deep, yin2019feature}. This tendency makes the linear classifier prone to favoring the dominant class, consequently causing the local models to exhibit bias after training. Considering the diverse decision boundaries among clients and distinct optimization directions for local models, challenges are posed to the generalization performance of aggregating the global model.}
	
	We propose a novel Federated Prototype Rectification with Personalization (FedPRP), which formulates personalized classifiers for heterogeneous clients with different skewed class distributions and improves representation learning through global consistency and robust optimization of estimated prototypes.
	Based on the decoupling principle of the model, the personalized classifier is a lightweight adaptive module tailored to the skewed class distributions of heterogeneous clients. Since it is optimized towards the client's own empirical risk, it ensures the personalized performance of local models.
	We design both inter-class discrimination and intra-class consistency losses that are applied to the representation to rectify the empirical prototypes derived from limited samples. 
	This guarantees that local training is more consistent across clients, making the aggregated global model more robust to skewed heterogeneous conditions and more generalizable to new clients.
	Besides, the global prototype on the server-side utilizes a moving average strategy for smooth updating, thereby mitigating the forgetting problem in the iterative training process.
	Our main contributions are summarized as follows:
	\begin{itemize}
		\item We briefly review the data heterogeneity of federated learning and further redefine the skewed heterogeneous federated learning, exploring how to realize dual goals that the generic performance or its personalized performance of the learned model.
		\item We propose a novel Federated Prototype Rectification with Personalization framework, which includes two modules: Federated personalization and Federated prototype rectification.
		\item Federated personalization entails the design of classifiers uniquely suited to each heterogeneous client, capable of training decision boundaries adapted to skewed class distributions, thereby ensuring the personalized performance of local models.
		\item To facilitate the training of generic high-quality representations and to regulate empirical prototypes, we introduce a federated prototype rectification based on inter-class discrimination and intra-class consistency.
	\end{itemize}
	\section{Related work}
	\noindent\textbf{Federated Learning:}
	Federated Averaging (FedAvg)~\cite{mcmahan2017communication} emphasizes that client training data is non-iid~\cite{sattler2020robust,2022Wei}, which is a common classic problem in the real world.
	{\color{black}For the non-iid data, Michieli et al.~\cite{michieli2021prototype} propose to use a prototypical representation of margin to learn distributed data to compute client-side biases and drive federated optimization through an attention mechanism.
		Based on the private data of each client, Fallah et al.~\cite{2020Personalized} investigate a personalized model variant of FL that provides each client with a more personalized model based on the meta-model through local training.
		To avoid the existence of data on the server-side, Zhu et al.~\cite{2021Data} propose a server-side knowledge distillation method without data for heterogeneous FL.
		For label distribution skew caused by non-iid, Zhang et al.~\cite{zhang2022federated} introduce a bias bound to measure the bias of the gradient after local updates and calibrate the logits before softmax cross-entropy according to the occurrence probability of each class.
		In addition, researchers have proposed some schemes based on cluster sampling, mainly according to the nature of the learning task or the similarity of the data samples~\cite{caldarola2021cluster,taik2022clustered,sattler2020clustered}.
		{\color{black}However, due to the lack of effective client selection strategies, the process of cluster sampling is relatively slow. Huang et al.~\cite{huang2023active} introduced active learning for the first time to select clients participating in each cluster, mainly based on some indicators of active learning that can provide the most information, and only aggregate their model updates to update specific model for clusters.}
		These methods focus on selecting representative clients and reducing the variance of randomly selected aggregation weights~\cite{2020An,2021Clustered,2020Briggs}.
		Although the abovementioned methods address data heterogeneity to some extent~\cite{bengio2013representation,lecun2015deep,collins2021exploiting,huang2022learn,sattler2020robust,zhu2019multi,shi2023towards}, they perform poorly on classes with few samples due to the lack of consideration of the long-tailed distribution.
	}

	\noindent\textbf{Skewed Heterogeneous Federated Learning:}
	In real-world scenarios, the datasets exhibit skewed distributions~\cite{kendall1946advanced, vannman2007process, zhang2023deep}. This means that a few dominant classes account for the majority of examples, while most other classes are represented by relatively few examples.
	Long-tailed distribution data has been widely studied in traditional centralized learning.
	Re-sampling method~\cite{junsomboon2017combining,hasib2022imbalanced,lin2023towards}: over-sampling or under-sampling for class samples, so that the number of samples of each class in the dataset is relatively balanced.
	Class weighting method~\cite{hazarika2022density, ganaie2022knn, guo2022learning}: assign a weight to each class in model training, so that the model pays more attention to learning minority classes.
	Data enhancement method~\cite{fu2022deep,wang2022gan}: perform data enhancement on samples of minority classes, such as rotation, flipping, etc., to increase the diversity of the dataset, thereby improving the generalization ability of the model.
	
	{\color{black}Existing research on long-tail learning typically assumes that all training samples are accessible during model training. However, in practical applications, skewed training data may be distributed across numerous heterogeneous mobile devices or the Internet of Things~\cite{luo2021no,sittijuk2021performance} and the data of these devices are non-iid. This constitutes a federated learning problem with data-level heterogeneity. Based on this observation, we propose skewed heterogeneous federated learning, which involves skewed class distributions, and these distributions may vary among different clients. This poses two key challenges: (1) imbalanced distribution of class data; (2) unknown shifts in class distribution among local data from different clients.
		
		To address these issues, Chou et al.~\cite{2021GRP} propose an adaptive aggregation parameter that adjusts the aggregation ratio to ensure fairness across clients.
		Shang et al.~\cite{2022Federated} construct a joint feature classifier to retrain heterogeneous and long-tail data and also introduce a new distillation method with logit adjustment and calibration gating network.
		Furthermore, Afonin et al.~\cite{afonin2021towards} introduce the ratio loss to estimate the global class distribution based on balanced proxy data on the server, allowing local optimization to achieve balance.
		However, this method necessitates relevant data on the server side for model training, which may compromise data security and prove to be impractical.
		The aforementioned methods all require relevant information about the data, which is not conducive to the realization of FL with data privacy protection.
		Our approach comprehensively takes into account the skewed FL problem that exists in real-world settings, where model bias and rapid initialization of new clients are systematically addressed without relying on proxy data.

	}
	\section{A Brief Review and the Proposed Setting}
	In this section, we briefly review existing work on data heterogeneity of federated learning and give a detailed and exact formulation for a more practical setting.
	\subsection{Data Heterogeneity of Federated Learning}
\begin{figure}[!htb]
		\centering
		\includegraphics[width=0.5\textwidth]{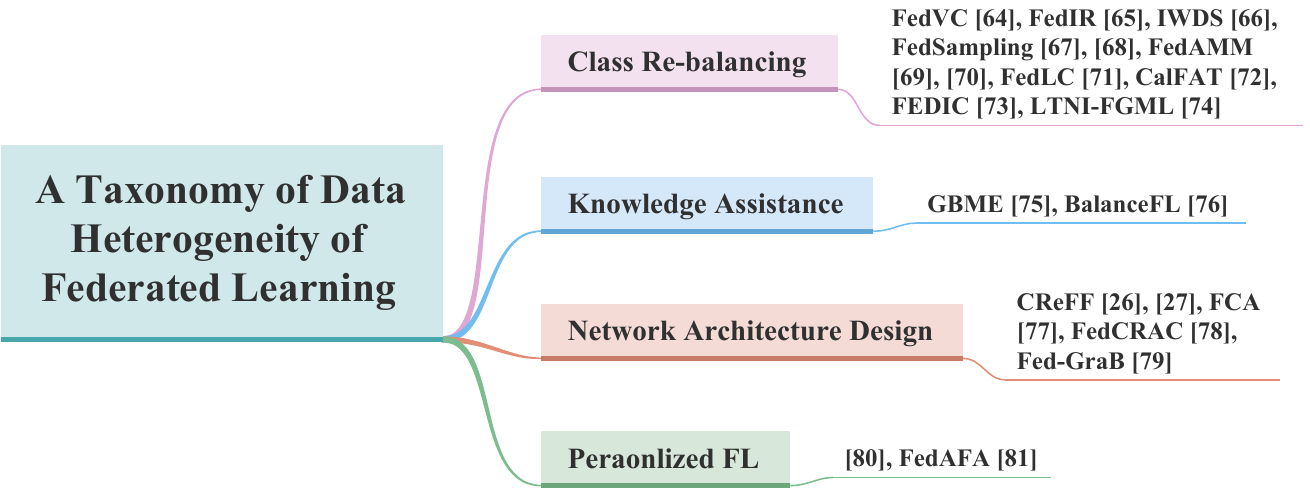}
		\caption{The taxonomy of data heterogeneity of federated learning.}
		\label{fig:2}
	\end{figure}

	Based on modeling strategies, we categorize existing methods into four types: class re-balancing methods, knowledge assistance methods, network architecture design methods, and personalized FL methods. The taxonomy is illustrated in~\Cref{fig:2}, and the details are summarized as follows.
	
	\noindent\textbf{Class Re-balancing} is a mainstream paradigm of SHFL that aims to mitigate the adverse effects of class imbalance without compromising privacy.
	The biased local model trained by the imbalanced data distribution within the client naturally affects the aggregation performance of the global model. To address this issue, training data re-sampling within clients~\cite{hsu2020federated,tang2021data,qi2023fedsampling,ding2023improved,qian2023long} and effective client selection~\cite{jiang2023secure} strategies have been explored.
	The method~\cite{zhang2023scoring} uses a scoring-based sampling strategy to select mobile clients with more few class data and better transmission conditions to upload their local models.
	Besides, some studies have confirmed the effectiveness of logit adjustment~\cite{wang2022logit,chen2022calfat,shang2022fedic,yan2023ltni} in solving data heterogeneity.
	
	\noindent\textbf{Knowledge Assistance} seeks to effectively use legitimate additional information within the FL framework to improve model performance for data heterogeneity. In contrast to centralized learning, the FL framework involves multiple client-side local models that collaboratively train a global model. The efficient use of the generated gradient information~\cite{zeng2023global} and other knowledge provides an interesting avenue to explore.
	The BalanceFL~\cite{shuai2022balancefl} applies a knowledge inheritance scheme to retain the knowledge in the global model for few-shot class supplementation to achieve sample balance.

	\noindent\textbf{Network Architecture Design} aims to improve various components of the network. Specifically, representation learning seeks to amplify the feature extractor; classifier design targets the enhancement of the model's classifier~\cite{FCA2023,yang2023integrating,li2023federated,xiao2023fed}; decoupled training~\cite{2022Federated} promotes parallel learning of the feature extractor and classifier; and ensemble learning seeks to refine the overall model framework.
	
	\noindent\textbf{Personalized FL} is regarded as a key element to solve such diverse data distribution problems for data heterogeneity.
	Reference~\cite{chen2022towards} indicates personalized FL as a promising technique that can improve the training performance of data heterogeneity through centralized long-tailed learning methods. Therefore, we study and discuss personalized FL as an independent type of method.
	FedAFA~\cite{lu2023personalized} optimizes each client's personalized model by generating a balanced feature set, thereby enhancing the capability for local minority classes.
	A series of studies{\color{black}~\cite{2020Personalized,deng2020adaptive,collins2021exploiting,2022Tan,li2023fedtp,liu2024federated}} have demonstrated that training personalized local models effectively resolves distributional differences.
	However, while personalized FL emphasizes client-specific nuances, it often neglects the generalization performance of the global model. Thus, the essence of personalized FL lies in striking an optimal balance between the globally shared representation and client-specific task modules to achieve true personalization.
	
	\subsection{Skewed Heterogeneous Federated Learning}
	There are several local clients $F = \{F_k\}_{k=1}^K$ that can participate in the training process, along with a central server $F_G$. The global model of $F_G$ is denoted as $\Theta$, while the local model of $F_k$ is denoted as $\theta_k$. The corresponding private data is written as $
	D_k = \left\{\left(\mathcal{X}_{k}, \mathcal{Y}_{k}\right) \mid \mathcal{X}_{k} \in \mathbb{R}^{M_{k} \times d},\\ \mathcal{Y}_{k} \in \mathbb{R}^{N \times 1}\right\}$, where $M_k$ indicates the number of instances, $d$ represents the input space, and $N$ is defined as the number of classes. This data distribution is indicated by $P_k(\mathcal{X},\mathcal{Y})$ and can be rewritten as $P_k(\mathcal{X}|\mathcal{Y}) \cdot P_k(\mathcal{Y})$.
	The Skewed Heterogeneous Federated Learning (SHFL) is represented as follows:
	\begin{itemize}
		\item $i, j \in [1, K], P_i(\mathcal{Y}) \neq P_j(\mathcal{Y})$.
		The label distribution $P(\mathcal{Y})$ across clients is inconsistent, i.e., there exists a class distribution shift across clients.
		\item$ i, j \in [1, N], \exists M_{i} \textless M_{j}$.
		There is a significant discrepancy between the overall sample sizes of the $i$-th class and the $j$-th class, i.e., there exists a skewed class distribution across clients from the perspective of the overall data.
		\item $k \in [1, K], i, j \in [1, N_k], \exists M_{k,i} \textless M_{k,j}$.
		The number of instances of the $i$-th and $j$-th classes is unequal within the $k$-th client, i.e., there exists a sample distribution skew within the client.
	\end{itemize}
	
	The SHFL inherently involves two major challenges: the skewness of local class distributions instigated by global class imbalances, and the unidentified class distribution shifts among local data from different clients. This results in pronounced disparities between local models, hindering their ability to maintain optimal consistency and subsequently affecting the aggregated global model's convergence to the genuine global optimum. Consequently, striking a balance between the personalization performance of the model and its generic capability is paramount. We emphasize private data distribution differences through personalization and improve generic representation through prototype rectification learning.
	{\color{black}
		
	}

	\section{The Proposed Method}
	In this section, we present a novel federated prototype rectification with personalization (FedPRP) framework to mitigate the challenges posed by SHFL. 
	The framework trains a private personalization module to adapt to its own class distribution and uses prototype rectification learning to encourage the model to learn high-quality representations of inter-class discrimination and intra-class consistency.
	
\begin{figure*}[!htb]
		\begin{center}
			\includegraphics[width=0.9\linewidth]{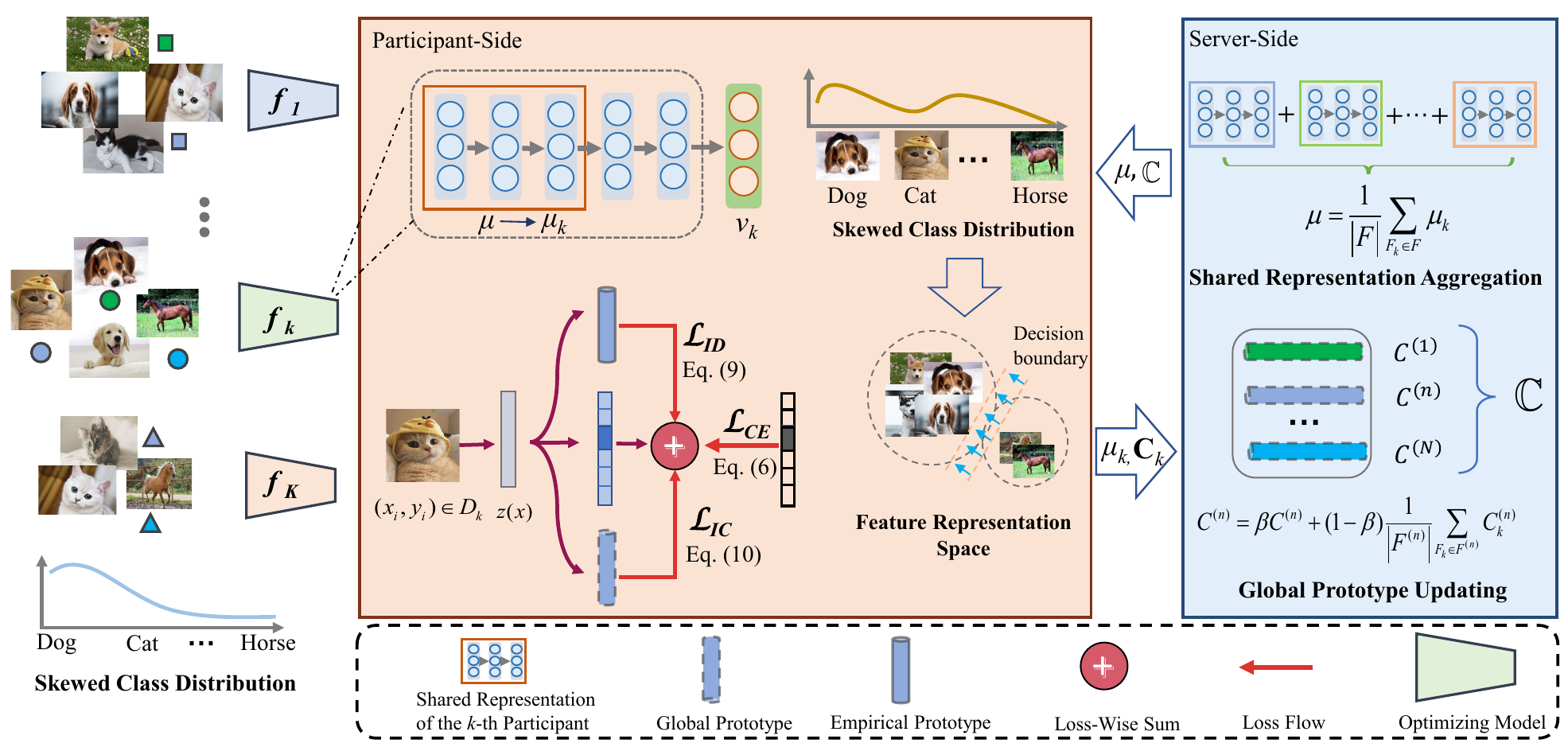}
		\end{center}
		\caption{{\color{black}Illustration of FedPRP. (a) Simplified schematization of our method that solves the SHFL issue via Federated personalization and Federated prototype rectification. 
				(b) Federated personalization: constructing the personalized module $v_k$ that adapts to unique decision boundaries for the class distribution shift. 
				(c) Federated prototype rectification: leveraging Inter-Class Discrimination Loss $\mathcal{L}^{k}_{\mathrm{ID}}$ and Intra-Class Consistency Loss $\mathcal{L}^{k}_{\mathrm{IC}}$ to learn a feature space that brings instances from the same class closer while pushing away those from different classes, ensuring consistency in global optimization to enhance the training of shared representations.}}
		\label{fig:overiew}
	\end{figure*}

	Figure~\ref{fig:overiew} illustrates an overview of FedPRP, which includes these key components: 
	(1) Federated personalization, which focuses on training individual private modules to fine-tune the unique decision boundary, addressing the class distribution shift across various clients; 
	(2) Federated prototype rectification, which uses the feature space of inter-class discrimination and intra-class consistency to improve the shared representation module.
	
	\subsection{Federated Personalization}
	In the personalized federated learning framework, we consider a local learning setting where each client $k$ trains its own model $\theta_k$ locally without communicating with other clients. The objective is given as:
	\begin{equation}
		\label{eq:per}
		\min _{\theta_{1}, \ldots, \theta_{k} \in \mathbb{R}^{d}} \mathcal{F}(\Theta):=\frac{1}{K} \sum_{k=1}^{K} \mathcal{F}_{k}\left(\theta_{k}\right),
	\end{equation}
	where $\theta_{k} \in \mathbb{R}^{d}$ encodes the parameters of the local model of client $k$. 
	A popular scheme is to decouple the local model into a shared representation module and personalization module, which is defined as:
	\begin{equation}
		\label{eq:split}
		\mathcal{F}_k\left(\theta_{k}\right) = f_{\mu_k}(\cdot)\circ g_{\nu_k}(\cdot),
	\end{equation}
	where $f_{\mu_k}(\cdot)$ denotes the shared representation module parameterized by $\mu_{k}$ and $g_{\nu_k}(\cdot)$ represents the personalized module parameterized by $\nu_{k}$.
	
	During local updating, the shared representation module is covered by the global $\mu$ for the gradient descent update:
	\begin{equation}
		\label{eq:share}
		\mu_{k}^{t+1} \gets \mu_{k}^{t} - \delta\frac{\partial \mathcal{F}_k\left(\mu^{t}, \nu_{k}\right)}{\partial x},
	\end{equation}
	where $\delta$ is the learning rate that controls the update step size, and $t$ denotes the number of rounds for updating.
	The multiple rounds of gradient descent are performed on the private personalization module to fit the unique class distribution and mitigate class distribution shift. The formulation can be defined as follows:
	\begin{equation}
		\label{eq:per}
		\nu_{k}^{s+1} \gets \nu_{k}^{s} - \delta\frac{\partial\mathcal{F}_k\left(\mu_{k}, \nu_{k}^{s}\right)}{\partial x},
	\end{equation}
	where $s$ is the number of rounds to update the personalized module, aims to train a private personalized decision boundary.
	
	The updated representation module is uploaded to the server for a round of global aggregation to obtain stronger generalization guarantees through cooperation among clients, which can be defined as:
	
	\begin{equation}
		\begin{split}
			\label{eq:agg}
			\mu = \frac{1}{|F|} \sum_{F_k \in F} \mu_{k},
		\end{split}
	\end{equation}
	where $F$ is the set of participants selected for training and $\mu_{k}$ is the shared low-dimensional representation module of $F_k$ after multiple local updates. 
	
	\noindent\textbf{Personalized Supervision Loss.}
	For local updating in federated learning, current methods typically cast this process as a supervised classification problem.
	Since the class distributions of the participants are shifted, each private model needs to adaptively fit a unique class distribution.
	Therefore, the personalized supervision loss of the $k$-th private model can be described as:
	\begin{equation}
		\label{CE}
		\mathcal{L}^k_{CE} =  -\frac{1}{|D_{k}|} \sum_{i=1}^{|D_{k}|} \sum_{j=1}^{N} \boldsymbol{y}_{ij} \log \mathcal{F}_k\left(\boldsymbol{x}_{i} ; \theta_k\right),
	\end{equation}
	where each $\left(\boldsymbol{x}_{i}, y_{i}\right) \in D_{k}$,  $\boldsymbol{y}_{ij}$ corresponds to the $j$-th element of one-hot encoded label of the sample $\boldsymbol{x}_{i}$, and $\boldsymbol{y}_{i}=\boldsymbol{e}_{y_{i}} \in\{0,1\}^{N}$ such that $\mathbf{1}^{\top} \boldsymbol{y}_{i} = 1 \forall i$.
	
	\subsection{Federated Prototype Rectification}
	Due to the limited number of training samples, the derived model may not achieve satisfactory generalization performance. To promote collaborative learning of the shared representation modules across clients, we employ a prototype communication scheme to transmit and aggregate prototype sets between the server and clients.
	
	Using the prototypes returned from the server, we employ a meticulously designed inter-class discrimination loss to incentivize the model to learn high-quality representations for both dominant and minority classes, thereby mitigating the bias of training empirical prototypes for limited samples. Concurrently, we perform local optimization using an intra-class consistency loss function that extracts class-related information, facilitating enhanced knowledge sharing in the latent space among participants.

	\noindent\textbf{Empirical Prototype:} To extract valuable class-related information from the local data, we construct the empirical prototype for knowledge uploading. Specifically, it is defined in the latent space of the projection network's output as the average of the representations within the same class $n$,
	\begin{equation}
		\label{eq:lp1}
		C^{(n)}_{k}:=\frac{1} {|D_k^{(n)}|} \sum_{(\boldsymbol{x},y) \in D_k^{(n)}} z\left(\boldsymbol{x}\right),
	\end{equation}
	where $z\left(\boldsymbol{x}\right)  = f_{u_k}\left(\boldsymbol{x}\right)$, $D_k^{(n)}$ indicates the subset of the private dataset $D_k$ that is comprised of training instances belong to the $n$-th class, and  $\mathbf{C}_k = \{C_k^{(n)}\}_{n=1}^{N}$ denotes the empirical prototype set of the $k$-th client.
	
	After the aforementioned computations, the empirical prototype set of participating clients is sent to the central server to update the global prototype using the weighted moving average:
	\begin{equation}
		\begin{split}
			\label{eq:GP1}
			\hat{C}^{(n)}= \beta \hat{C}^{(n)} + (1-\beta) \frac{1}{\left|F^{(n)}\right|} \sum_{F_k \in F^{(n)}} C^{(n)}_k,
		\end{split}
	\end{equation}
	where $F^{(n)}$ denotes the set of clients that own instances of class $n$, and $\beta$ refers to the decay rate of the moving average model, which can be used to control the iterative update rate of the prototype.
	The moving average approach has the advantage of smoothing updates in the new sample data to update the old class prototypes, thus making them more generalizable. When a new class prototype is added, the value of $\beta$ is set to $0$.
	The global prototype set $\mathbb{C} = \{\hat{C}^{(n)}\}_{n=1}^{N}$ can be obtained by iterative updates on the server-side.
	
	Since the averaging is an irreversible process~\cite{tan2022fedproto} it is impossible to reconstruct the original data from the prototype if the attacker does not have access to the local model data, passing of the prototype does not invalidate the privacy protection.
	
	\noindent\textbf{Inter-Class Discrimination Loss:} 
	Following empirical risk minimization, we can naively obtain the empirical prototype $C^{(n)}_{k}$ through~\cref{eq:lp1}. 
	However, due to limited training data within the client, the true prototype is unknown and its estimation is noisy. We hope to design a robust loss that can pull samples closer to the centroid of their own class while pushing away samples from other classes.
	
	Given the distributionally robust optimization~\cite{samuel2021distributional}, which defines a set of uncertain test distributions around the training distribution and learns a classifier to minimize the expected risk under the worst-case test distribution.
	Inspired by this, we construct an inter-class discrimination loss for representation that extends the standard contrastive loss, defined as
	
	\begin{equation}
		\label{ID}
		\mathcal{L}^k_{\text {ID}}=-\sum_{n \in N} \sum_{(\boldsymbol{x},y) \in D_k^{(n)}} \log \frac{e^{-d\left(C^{(n)}_{k}, z_{\boldsymbol{x}}\right)-\epsilon }}{\sum_{z_{\boldsymbol{x}}^{\prime}} e^{-d\left(C^{(n)}_{k}, z_{\boldsymbol{x}}^{\prime}\right)- \epsilon^{\prime} }},
	\end{equation}
	where $z_{\boldsymbol{x}}$ represents $z(\boldsymbol{x})$ for short, $d\left(C^{(n)}_{k}, z_{\boldsymbol{x}}\right)$ measures the distance in feature space between $z_{\boldsymbol{x}}$ and the empirical prototype of its class, $\epsilon $ and $\epsilon^{\prime}$ are robustness margins. Specifically, we consider here the case where $d$ is the Kullback-Leibler divergence and treat the $\epsilon$ as a learnable parameter, adjusting its value during training.

	\noindent\textbf{Intra-Class Consistency Loss:}
	{\color{black}To enforce the representation $z_{\boldsymbol{x}}$ generated by the local projection network to align with its own global class centers, learning more class-related but client-irrelevant information.
		We introduce an intra-class consistency loss based on the global prototype, defined as:
		\begin{equation}
			\label{IC}
			\mathcal{L}^{k}_{\mathrm{IC}} = \frac{1}{|D_{k}|}\sum_{n \in N} \sum_{(\boldsymbol{x}, y) \in D^{(n)}_{k}} \left| z_{\boldsymbol{x}} - \hat{C}^{(n)} \right|^2_2,
		\end{equation}
		where the global prototype $\hat{C}^{(n)}\in \mathbb{C}$. The empirical prototypes optimized using the global prototype regularization do not deviate from the center of the class, which also avoids significant deviations during subsequent iterations of global prototype updates.
	}
	
	\subsection{Training and Inference}
	The personalized supervision loss in~\cref{CE} and the inter-class discrimination loss in~\cref{ID} with intra-class consistency loss in~\cref{IC} synergistically enhance each other.
	The former implements private personalized classification to alleviate cross-client distribution shift, while the latter two implement the enhancement of shared representation modules from an inter-class perspective to improve generalization.
	The overall training target for the $k$-th local model is:
	\begin{equation}
		\begin{split}
			\label{eq:taotal}
			\mathcal{L}^{k}_{\mathrm{Total}}  = \mathcal{L}^{k}_{CE}+ \lambda \cdot \mathcal{L}^{k}_{\mathrm{ID}} + (1-\lambda) \cdot \mathcal{L}^{k}_{\mathrm{IC}}.
		\end{split}
	\end{equation}
	
	By optimizing \cref{eq:taotal}, the generalization capability of the shared representation module is enhanced. The parameters of this module are then uploaded to the server along with the empirical prototype set for aggregation to facilitate collaborative learning.
	The global prototype is updated via a moving average as given in~\cref{eq:GP1} to mitigate knowledge forgetting during the training process.
	
	{\color{black}
		We have provided a detailed description of the FedPRP algorithm in ~\cref{algorithm: AllModel}. The central server initializes a set of category prototypes and collaborates with clients during iterative training. In each round, clients utilize $\mathcal{L}^k_{\text{ID}}$ and $\mathcal{L}^k_ {\text{IC}}$ for ensuring the consistency of category prototypes. The global prototypes are updated based on the aggregated information from local updates, and the shared representation module is refined to capture collective knowledge. Lines 1 to 9 describe the aggregated updates made to the server-side global model and global prototypes, while lines 11 to 19 outline the local update process performed on the client-side.}
	
	During the validation phase of the $F_k$, we use the distance between the embedding representation of the test sample $\mathbf{\hat{x}}$ and the existing class prototype within the client to obtain the prediction label $\hat{y}$. The process can be described as follows:
	\begin{equation}
		\begin{split}
			\label{eq:test}
			\hat{y}=\underset{n}{\arg \min }\left\| f_{\mu_k}\left(\mathbf{\hat{\boldsymbol{x}}}\right) - C_k^{(n)}\right\|_{2}, \mathbf{\hat{\boldsymbol{x}}} \in \mathcal{X}^{Test},
		\end{split}
	\end{equation}
	where the $f_{\mu_k}\left(\mathbf{\hat{\boldsymbol{x}}}\right)$ function gets the embedding vector of $\mathbf{\hat{\boldsymbol{x}}}$, $\mathcal{X}^{Test}$ denotes the total test instances and $C_k^{(n)}$ is the empirical prototype on the $k$-th client. If we use the server-side global prototype for prediction then just replace it with $\hat{C}^{(n)}$.
	%
	\begin{algorithm}[!htb]
		\caption{FedPRP}\label{algorithm: AllModel}
		\hspace*{0.02in}{\bf Input:} {$D_k, \theta_{k},k = 1,2,\dots$, and a pre-trained global shared representation module $\mu$.} \\
		\hspace*{0.02in}{\bf Output:} {$\mu_k$ and $\mathbf{C}_k$.}
		\begin{algorithmic}[1]
			\STATE \textbf{Central Server:}
			\STATE Initialize prototype set $\{C^{(n)}\}_{n=1}^{N}$;
			\FOR{each round $T = 1,2,\dots $}
			\FOR{each client $k$ \textbf{in parallel}}
			\STATE $\mu_k, \mathbf{C}_k \gets$ LocalUpdate $(k, \mu, \mathbb{C}, \{C^{(n)}\}_{n=1}^{N})$;
			\ENDFOR
			\STATE Update global prototype by~\cref{eq:GP1};
			\STATE Update global shared representation module $\mu$ by~\cref{eq:agg};
			\ENDFOR
			\STATE \textbf{LocalUpdate($k,\mu, \mathbb{C}$):}\\
			\FOR{each local epoch}
			\FOR{each batch in $D_k$}
			\STATE Compute $\mathcal{L}^k_{\text {ID}}$ by~\cref{ID} with empirical prototypes;
			\STATE Compute $\mathcal{L}^k_{\text {IC}}$ by~\cref{IC} with global prototypes;
			\STATE Update $\theta_k$ by~\cref{eq:taotal};\\
			\STATE $(\mu_k;\nu_k)\gets \theta_k$;
			\ENDFOR
			\ENDFOR
			\STATE Compute empirical prototypes by~\cref{eq:lp1};
		\end{algorithmic}
	\end{algorithm}
	
	\section{Experiments}
	In this section, we experimentally verify the effectiveness of the FedPRP by comparing it with representative methods. Further experiments analyzing the generalization performance and ablation studies are also provided.
	
	\subsection{Experimental Settings}\label{section:exset}
	\subsubsection{Datasets}
	We utilize three popular benchmark datasets: CIFAR10, CIFAR100~\cite{krizhevsky2009learning}, and Tiny-ImageNet.
	\begin{itemize}
		\item CIFAR10~\cite{krizhevsky2009learning} dataset is a popular benchmark for image classification tasks. It consists of 60,000 32 $\times$ 32 color images across ten classes, with 6,000 images per class.
		\item CIFAR100~\cite{krizhevsky2009learning} is an extension of the CIFAR10 dataset and is designed for more fine-grained image classification. It contains 100 classes, each with 600 images.
		\item Tiny-ImageNet~\cite{le2015tiny} dataset is a subset of the ILSVRC2012~\cite{krizhevsky2012imagenet} classification dataset. It consists of 200 object classes, and each object class contains 500 training images, 50 validation images, and 50 test images. All images have been downsampled to dimensions of 64 $\times$ 64 $\times$ 3 pixels.
	\end{itemize}
	
	To establish a heterogeneous data environment with globally skewed class distributions across clients, we define the imbalance ratio ($\gamma$) as the ratio between the sample count of the least frequent class and the most frequent class as the described in~\cite{cao2019learning}, i.e., $\gamma =\min_{i}\left\{n_{i}\right\}/\max_{i}\left\{n_{i}\right\}$. We set $\gamma \in \{0.05, 0.1, 0.5, 1\}$.
	Moreover, the inherent heterogeneity among client data leads to disparate class distributions, which we simulate using two data partitioning strategies:
	\begin{itemize}
		\item \textit{Sharding}~\cite{collins2021exploiting}: sorts all training sample data by label, divides the data into equal-sized shards based on the total number of clients, and controls heterogeneity via $s$ (the number of classes in each client). The data size of each client is the same, but the class distribution is not uniform. Specifically, we set $s$ to CIFAR10 ($s$ = 4, 5), CIFAR100 ($s$ = 20, 30), and Tiny-ImageNet ($s$ = 20). 
		\item \textit{Dirichlet Distribution Allocation (DDA)}~\cite{luo2021no,lee2022preservation}: assigns a partition of class $k$ by sampling $P_k \approx Dir(\alpha)$ according to the Dirichlet distribution. The value of $\alpha$ controls how skewed the label distribution is, with larger values implying less variance in label distribution between clients. The class distribution and data size are different among clients in \textit{DDA}. Specifically, we set $\alpha \in \{0.05, 0.4\}$ on experimental to investigate different scenarios.
	\end{itemize}
	{\color{black} The selection of $s$ and $\alpha$ aligns with the prevailing approaches in the existing literature~\cite{collins2021exploiting,tan2022fedproto}. The constraint for selecting $s$ lies in its limitation not to exceed the number of classes in the dataset. To manifest noticeable heterogeneity, a smaller value can be assigned to $\alpha$, while a comparison analysis has been conducted with the selection of larger $\alpha$ values.}
	
	\subsubsection{Implement Details}
	Initially, when the data distribution is highly imbalanced $\gamma = 0.05$, we set up 20 local clients ($K = 20$) and randomly selected 8 client participants in each round.
	Additionally, to increase the degree of heterogeneity, we set up 100 clients for other imbalanced situations ($\gamma \in \{0.1, 0.5, 1\}$), and the central server randomly selected 10 of the clients ($KS = 10$) to participate in the learning process.
	For all methods, we set the local training epoch to $t+s = 30$ and the global training epoch to 100.
	We employ the ResNet18~\cite{He2016Deep} as the basic backbone network for local models of all participants, with a learning rate of 0.01 and the SGD optimizer.
	Furthermore, we set $\beta$ = 0.5 to balance the knowledge of old and new samples for the existing class, and $\lambda$ = 0.5 to weigh the inter-class discrimination loss and intra-class consistency loss.
	
	\subsubsection{Evaluation Metrics}
	We report the standard metric of method quality: accuracy, which is defined as the number of correctly predicted samples divided by the total number of samples.
	To evaluate the performance of the local and the global models separately, we define the following three different metrics.
	
	$\mathcal{A}^{glo}$ is the prediction accuracy of the balanced test set on the aggregated global model.
	\begin{equation}
		\begin{split}
			\mathcal{A}^{glo}=\frac{\sum\left(\operatorname{argMax}\left(\mathcal{F}_\Theta\left(\mathcal{X}^{Test}\right)\right)==\mathcal{Y}\right)}{\left|D^{Test}\right|}.
			\nonumber
		\end{split}
	\end{equation}
	$\mathcal{A}^{loc}$ is the prediction accuracy of the balanced test set on the private personalized model.
	\begin{equation}
		\begin{split}
			\mathcal{A}^{loc}=\frac{1}{KS}\sum_{k\in KS}\frac{\sum\left(\operatorname{argMax}\left(\mathcal{F}_{\theta_k}\left(\mathcal{X}^{Test}\right)\right)==\mathcal{Y}\right)}{\left|D^{Test}\right|},
			\nonumber
		\end{split}
	\end{equation}
	where $\mathcal{F}_{\theta_k}$ represents $\mathcal{F}_{k}(\theta_k)$ for short, $D^{Test}$ is the balanced test set of all classes, and $\mathcal{Y}$ is the label set of the total testing instances.
	$\Theta$ is the global model and $\theta_k$ is the local model of the $k$-th client.
	$\mathcal{A}_{k}^{sel}$ is the prediction accuracy on the test set of the $k$-th client, partitioned by the class distribution corresponding to the training phase, using the private personalized model.
	
	\begin{equation}
		\begin{split}
			\label{eq:share}
			\mathcal{A}_{k}^{sel}=\frac{\sum\left(\operatorname{argMax}\left(\mathcal{F}_{\theta_k}\left(\mathcal{X}_k^{Test}\right)\right)==\mathcal{Y}\right)}{\left|D_k\right|}.
			\nonumber
		\end{split}
	\end{equation}
	
	For all datasets, we further report accuracy on three splits of the set of classes. ``\textit{Many}": classes with the top 20\% number of training samples; ``\textit{Medium}": classes with the top 20\% - 50\% number of training samples; and ``\textit{Few}": classes with the bottom 50\% number of training samples.

	\subsubsection{Baselines} We compare the proposed method with the popular state-of-the-art heterogeneity FL methods, i.e., FedAvg~\cite{mcmahan2017communication,collins2021exploiting}, FedRep~\cite{collins2021exploiting}, FedProx~\cite{li2020federated}, APFL~\cite{deng2020adaptive}, FedRod~\cite{chen2021bridging}, FedPHP~\cite{li2021fedphp} and FedProto~\cite{tan2022fedproto}.
	We also select two representative FL methods that address the long-tailed distribution, i.e., FEDIC~\cite{shang2022fedic} and CReFF~\cite{2022Federated}.
	CReFF~\cite{2022Federated} extends the idea of decoupling the training process to retrain classifiers based on joint features in a privacy-preserving manner, while FEDIC~\cite{shang2022fedic} is a method that uses logit to adjust and calibrate the gate network.
	{\color{black}FedRod~\cite{chen2021bridging} achieves both general and personalized performance of the learning model by introducing a series of losses that are robust to different class distributions and formulating lightweight adaptive modules. FedPHP~\cite{li2021fedphp} explores the use of temporary sets of historical personalization models to supervise the next round of global personalization processes.}
	All baseline methods were performed with the same data settings.
	
	\subsection{Experimental Analysis}\label{section:Analysis}
	\subsubsection{Comparison with State-of-the-art Methods}
	We compare the proposed method with state-of-the-art federated learning methods in various scenario settings. Table~\ref{tab:examplesh} reports the average $\mathcal{A}^{loc}$ for the participants in the last ten rounds under the \textit{Sharding} non-iid partitioning strategy.
	
	\begin{table*}[!htb]\footnotesize
		\centering
		{\color{black}
			\caption{The $\mathcal{A}^{loc}$ (\%) of different FL methods on all datasets under the \textit{Sharding} non-iid partition strategy. Note that we highlight the \textbf{best} results in bold and the \underline{second best} results in underlining.}
			\resizebox{1\linewidth}{!}{
				\begin{tabular}{c|ccc|ccc|ccc|ccc|c}
					\hline
					Datasets & \multicolumn{6}{c|}{CIFAR10}& \multicolumn{6}{c|}{CIFAR100}& \multicolumn{1}{c}{Tiny-ImageNet}\\
					\midrule
					\textit{Sharding}& \multicolumn{3}{c|}{$s$ =  4}&\multicolumn{3}{c|}{$s$ =  5}&\multicolumn{3}{c|}{$s$ =  20}&\multicolumn{3}{c|}{$s$ =  30}& \multicolumn{1}{c}{$s$ =  20}\\
					$\gamma$ & 0.1 & 0.5 &  1.0 &  0.1 & 0.5 &  1.0 &  0.1 & 0.5 &  1.0 & 0.1 & 0.5 &  1.0 & 0.5 \\
					\midrule
					APFL~\cite{deng2020adaptive} &46.11&54.74&54.93&48.54&62.17&63.54&19.95&25.92&29.54&19.37&26.21&30.13&18.31  \\
					FedRep~\cite{collins2021exploiting} &46.87&58.44&63.44&50.51&61.57&68.65&24.04&29.87&33.87&23.31&30.23&33.87&21.91 \\
					FedProx~\cite{li2020federated} &52.87&\underline{61.47}&63.48&51.78&\underline{67.35}&\underline{69.41}&22.81&29.40&32.94&22.18&29.06&32.67&28.19  \\
					FedProto~\cite{tan2022fedproto} &49.84&58.54&62.57&45.19&52.81&56.04&21.65&30.01&33.79&16.62&21.91&25.42&29.55  \\
					FedRod~\cite{chen2021bridging}&45.97&59.85&\underline{64.99}&47.30&60.75&64.78&23.77&30.73&35.74&23.95&30.87&34.32&29.15\\
					FedPHP~\cite{li2021fedphp}&48.38&56.94&61.44&48.27&64.01&68.28&20.62&27.82&30.03&20.83&27.82&30.22&\underline{30.16}\\
					FedAvg~\cite{mcmahan2017communication} &43.03&53.08&56.90&44.79&56.89&62.11&23.15&30.24&33.81&30.01&30.32&33.53&27.68\\
					\midrule
					FEDIC~\cite{shang2022fedic} & \underline{60.11} & 61.29 & 60.94 & \underline{60.84} &	60.52&64.05&\underline{31.07}&32.52&33.32&\textbf{30.13}&32.11&32.58&29.57 \\
					CReFF~\cite{2022Federated} &47.56&56.88&60.24&49.40&61.58&64.94&28.41&\underline{35.64}&\underline{41.01}&26.78&\underline{34.97}&\underline{39.52}&27.99  \\
					\midrule
					FedPRP&\textbf{74.28}&\textbf{81.07}&\textbf{83.74}&\textbf{70.74}&\textbf{79.73}&\textbf{82.58}&\textbf{35.46}&\textbf{43.52}&\textbf{48.12}&\underline{28.62}&\textbf{37.59}&\textbf{41.22}&\textbf{40.36}\\
					\hline
			\end{tabular}}
			\label{tab:examplesh}}
	\end{table*}
	The performance of the FedPRP method has consistently delivered competitive results across various datasets with diverse settings. Especially when the $s$ = 4 and $\gamma$ = 0.1 on the CIFAR10 dataset, our proposed approach outperforms the suboptimal method by a significant margin of 14.17\%.
	Furthermore, the FEDIC and CReFF methods, which are designed to address the challenge of long-tailed heterogeneity, exhibit substantial advantages when applied to the CIFAR100. Conversely, the personalized methods demonstrate exceptional performance on CIFAR10. This observation suggests that the skewed long-tailed problem becomes more prominent as the number of classes increases.
	On the Tiny-ImageNet dataset, FedPRP outperforms FedProto by 10.81\% over prototype learning, providing strong evidence that prototype knowledge can aid in learning high-quality representations.

	\begin{table}[!htb]\footnotesize
		\centering{{\color{black}
				\caption{The $\mathcal{A}^{glo}$ (\%) of different FL methods on CIFAR10 dataset under the \textit{DDA} non-iid partition strategy.}
				\resizebox{1\linewidth}{!}{
					\begin{tabular*}{9cm}{@{\extracolsep{\fill}}cccc|ccc}
						\toprule
						&\multicolumn{3}{c}{$\alpha$ = 0.4}&\multicolumn{3}{c}{$\alpha$ = 0.05}\\
						\midrule
						$\gamma$& 0.1 & 0.5 &1.0&0.1 & 0.5 &1.0 \\
						\midrule
						APFL~\cite{deng2020adaptive} &43.27 & 57.51 & 63.30 & 47.43 & 45.88 & 50.45\\
						FedRep~\cite{collins2021exploiting} &44.48 & 56.10 & 63.27 & 39.01 & 48.43 & 52.23\\
						FedProx~\cite{li2020federated} & 43.26& 56.21 & 62.83 & 38.23 & 45.39 & \underline{62.83}\\
						FedProto~\cite{tan2022fedproto} & 32.33 & 34.63 & 34.16 & 47.78 & 45.07 & 47.42\\
						
						FedRod~\cite{chen2021bridging}&51.60&60.79&65.07&47.89&53.45&56.84\\
						FedPHP~\cite{li2021fedphp}&49.51&53.71&64.42&43.67&51.07&62.79\\
						FedAvg~\cite{mcmahan2017communication} &43.44 & 55.57 & 63.02 & 38.19 & 45.19 & 50.35\\
						\midrule
						FEDIC~\cite{shang2022fedic} &  \underline{57.01} & 59.02 & 59.16 & \underline{58.99} & \underline{59.79} & 57.74  \\
						CReFF~\cite{2022Federated} &53.75 & \underline{61.79} & \underline{65.94} & 49.51 & 52.17 & 54.63\\
						\midrule
						FedPRP&\textbf{60.71}&\textbf{66.82}&\textbf{71.94}&\textbf{74.39} &\textbf{75.25}&\textbf{79.19}\\
						\bottomrule
				\end{tabular*}}
				\label{exampledir}}}
	\end{table}
	Table~\ref{exampledir} lists the average results of the last ten rounds of $\mathcal{A}^{glo}$ on the CIFAR10 dataset for different imbalance ratios. The proposed method achieves excellent results consistent with the \textit{Sharding} non-iid partitioning strategy and has advantageous improvements on different datasets.
	
	\begin{table*}[!htb]
		\centering
		\caption{The $\mathcal{A}^{glo}$ (\%) for FedPRP and compared FL methods on CIFAR10 with $\gamma$ = 0.05.}
		\resizebox{0.9\linewidth}{!}{
			\begin{tabular}{c c c c c c | c c c c c }
				\toprule
				& \multicolumn{10}{c}{CIFAR10}\\
				& \multicolumn{5}{c}{\textit{Sharding}: $s$ =  4}& \multicolumn{5}{c}{\textit{DDA}: $\alpha$ = 0.4}\\
				\cmidrule{2-11}
				& \textit{\textbf{All}} & Many&  Medium & Few  & \textit{\textbf{Average}} & \textit{\textbf{All}} &  Many&  Medium & Few& \textit{\textbf{Average}} \\
				\midrule
				APFL~\cite{deng2020adaptive}     & 45.62 & 40.93 & 36.68 & 35.71 & 37.77 & 49.65 & 91.08 & 47.14 & 25.83 & 54.68 \\
				FedRep~\cite{collins2021exploiting}   & 50.77 & 46.77 & 38.83 & 39.94 & 41.85 & 50.02 & 91.37 & 48.55 & 25.16 & 55.03 \\
				FedProx~\cite{li2020federated}  & 47.59 & 41.42 & 37.95 & 37.85 & 39.07 & 50.22 & 90.44 & 48.38 & 25.88 & 54.90 \\
				FedProto~\cite{tan2022fedproto} & \underline{65.41} &	54.94 & 59.03 & 46.13  &\underline{53.37} & 48.36 & 73.41 & 41.18 & 18.79 & 44.46 \\
				FedAvg~\cite{mcmahan2017communication}  & 46.19 & 40.81 & 37.62 & 36.26 & 38.23 & 50.55 & 91.14 & 48.33 & 24.70  & 54.72 \\
				\midrule
				FEDIC~\cite{shang2022fedic}& 63.94 & 44.48 & 44.63 & 64.57 & 51.23& 61.99 & 69.36 & 45.15 & 45.43 & 53.31 \\
				CReFF~\cite{2022Federated}    & 56.53 & 47.11 & 48.52 & 43.63 & 46.42 & \underline{63.86} & 90.63 & 54.67 & 31.74 & \underline{59.01} \\
				\midrule
				FedPRP& \textbf{74.23} &	56.09 &	64.01 &	59.83 &\textbf{	59.98} &\textbf{66.26} &	91.48 &	55.74 	&34.09& \textbf{60.44 }\\
				\bottomrule
		\end{tabular}}
		\label{tab:cifar10}
	\end{table*}
	
	We set up a total of 100 clients to simulate large-scale complex application scenarios, and the proposed method stands out from other research methods. Especially when $\alpha$ = 0.05 and $\gamma$ = 0.1 have strong skew heterogeneity, FedPRP is 15.40\% higher than the FEDIC method.
	It can be seen that in complex scenarios of skew heterogeneous federated learning, the effectiveness of the FEDIC and CReFF methods is not particularly obvious.

\begin{figure*}[!htb]
		\begin{center}
			\includegraphics[width=0.65\linewidth]{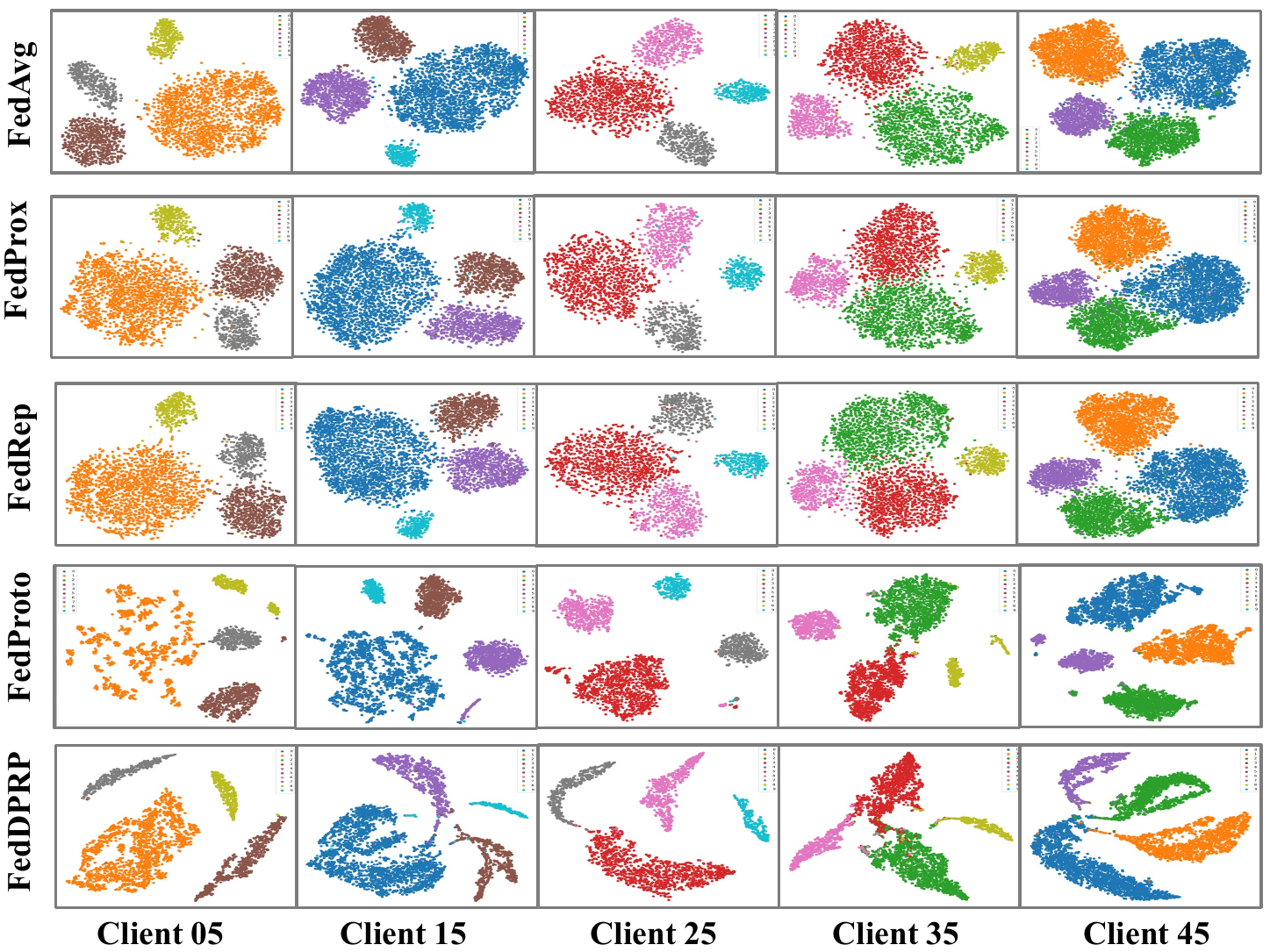}
			\caption{{\color{black}t-SNE feature visualization of features learned with different methods on $\gamma$ = 0.1 and $s$ = 4. Inconsistent class distribution among different participants, and skewed sample distributions within clients. Features are colored based on classes.}}
			\label{fig:tsne}
		\end{center}
	\end{figure*}

	{\color{black}\subsubsection{Feature Visualizations}
		To elucidate the impact of skew heterogeneity on feature representation capabilities, we employ t-distributed Stochastic Neighbor Embedding (t-SNE)~\cite{van2008visualizing} to visualize representations trained from local models. \Cref{fig:tsne} illustrates t-SNE feature visualizations on five different clients under different federated learning algorithms.}
	
	{\color{black}Obviously, after several epochs of training, the features acquired by FedPRP become more compact and show improved separability both within and between classes. This improvement can be attributed to the high quality representations learned via joint prototype refinement, thus improving the classification representation of personalized modules. Even for classes with limited samples, promising intra- and inter-class performance is demonstrated. Among them, prototype-based FedProto also shows good separability performance, but the biased number of class prototypes may lead to misclassification. In addition, other federated optimization methods manifest relatively fuzzy class distinctions.}
	
	
	\subsubsection{Accuracy for Different Groups of Classes}
	We divide the classes into three groups based on the number of samples: \textit{Many}, \textit{Medium}, and \textit{Few}.
	Some methods for addressing the long-tailed problem often prioritize improving the accuracy of the \textit{Few} class, potentially at the expense of accuracy for the \textit{Many} and \textit{Medium} classes. However, it is important to note that these two groups of classes are common in real-world scenarios, making precise predictions for both of paramount importance. For this purpose, we use the notation \textit{\textbf{All}} to represent the prediction accuracy of all samples within a balanced test set, while \textit{\textbf{Average}} denotes the average prediction accuracy across the three groups.
	Tables~\ref{tab:cifar10} and~\ref{tab:splitcifar100} present the experimental results obtained using various non-iid partition strategies on the CIFAR10 and CIFAR100 datasets, respectively.
	
	As observed in~\Cref{tab:cifar10}, our proposed methodology effectively improves the accuracy of the \textit{Few} class while maintaining the accuracy levels of the \textit{Many} and \textit{Medium} classes.
	More specifically, when the \textit{Sharding} strategy is employed, FEDIC achieves remarkable results for the \textit{Few} class; however, it falls short of the suboptimal FedProto method by approximately 10.46\% in terms of accuracy for the \textit{Many} class. Moreover, the accuracy for the \textit{Medium} class experiences a decrease of about 14.40\%.
	Although our proposed method is lower than FEDIC in the \textit{Few} class, it is better than other personalization methods by about 13.7\% to 24.12\% and has the best effect on \textit{\textbf{Average}}, which is higher than FEDIC 6.61\%.
	
	\begin{table*}[!htb]
		\centering
		\caption{The $\mathcal{A}^{glo}$ (\%) for FedPRP and compared FL methods on CIFAR100 with $\gamma$ = 0.05.}
		\resizebox{0.9\linewidth}{!}{
			\begin{tabular}{c c c c c c| c c c c c}
				\toprule
				& \multicolumn{10}{c}{CIFAR100}\\
				& \multicolumn{5}{c}{\textit{Sharding}: $s$ =  20}& \multicolumn{5}{c}{\textit{DDA}: $\alpha$ = 0.4}\\
				\cmidrule{2-11}
				& \textit{\textbf{All}} & Many&  Medium & Few  & \textit{\textbf{Average}} & \textit{\textbf{All}} &  Many&  Medium & Few& \textit{\textbf{Average}} \\
				\midrule
				APFL~\cite{deng2020adaptive} &25.51 & 37.42 & 29.97 & 17.23 & 28.21    & 26.02 & 41.85 & 31.81 & 16.22 & 29.96    \\
				FedRep~\cite{collins2021exploiting} &27.07 & 36.99 & 30.82 & 19.62 & 29.15    & 26.42 & 42.13 & 31.85 & 16.99 & 30.32    \\
				FedProx~\cite{li2020federated} &25.37 & 37.96 & 29.81 & 16.96 & 28.24    & 26.01 & 42.25 & 30.55 & 16.69 & 29.83    \\
				FedProto~\cite{tan2022fedproto} &33.06 & 55.54 & 38.87 & 20.05 & 38.15    & 28.57  & 38.60 & 20.08 & 15.01  & 24.56    \\
				FedAvg~\cite{mcmahan2017communication} &25.45 & 37.48 & 29.21 & 17.37 & 28.02    & 26.05 & 42.48 & 31.07 & 16.11 & 29.89    \\
				\midrule
				FEDIC~\cite{shang2022fedic}&\underline{35.50} &	61.27 &	35.79 &	25.09 &	\underline{40.72}&	
				\underline{32.89} & 35.89 & 32.39 & 25.85 & \underline{31.38} \\
				CReFF~\cite{2022Federated} &33.38& 49.95 & 43.37 & 27.12 & 40.15&
				32.24 & 41.85 & 34.12 & 17.15 & 31.04 \\
				\midrule
				FedPRP& \textbf{40.88} & 62.99 &	47.47& 	27.63&\textbf{46.03 }&\textbf{35.43} & 45.63 & 35.31 & 18.90 & \textbf{33.28}\\
				
				\bottomrule
		\end{tabular}}
		\label{tab:splitcifar100}
	\end{table*}
	
	\begin{figure*}[!tb]\footnotesize
		\centering{\color{black}
			\subfloat[$s$ = 4, CIFAR10 ]{\label{figure_ca}\includegraphics[height=0.2\textwidth]{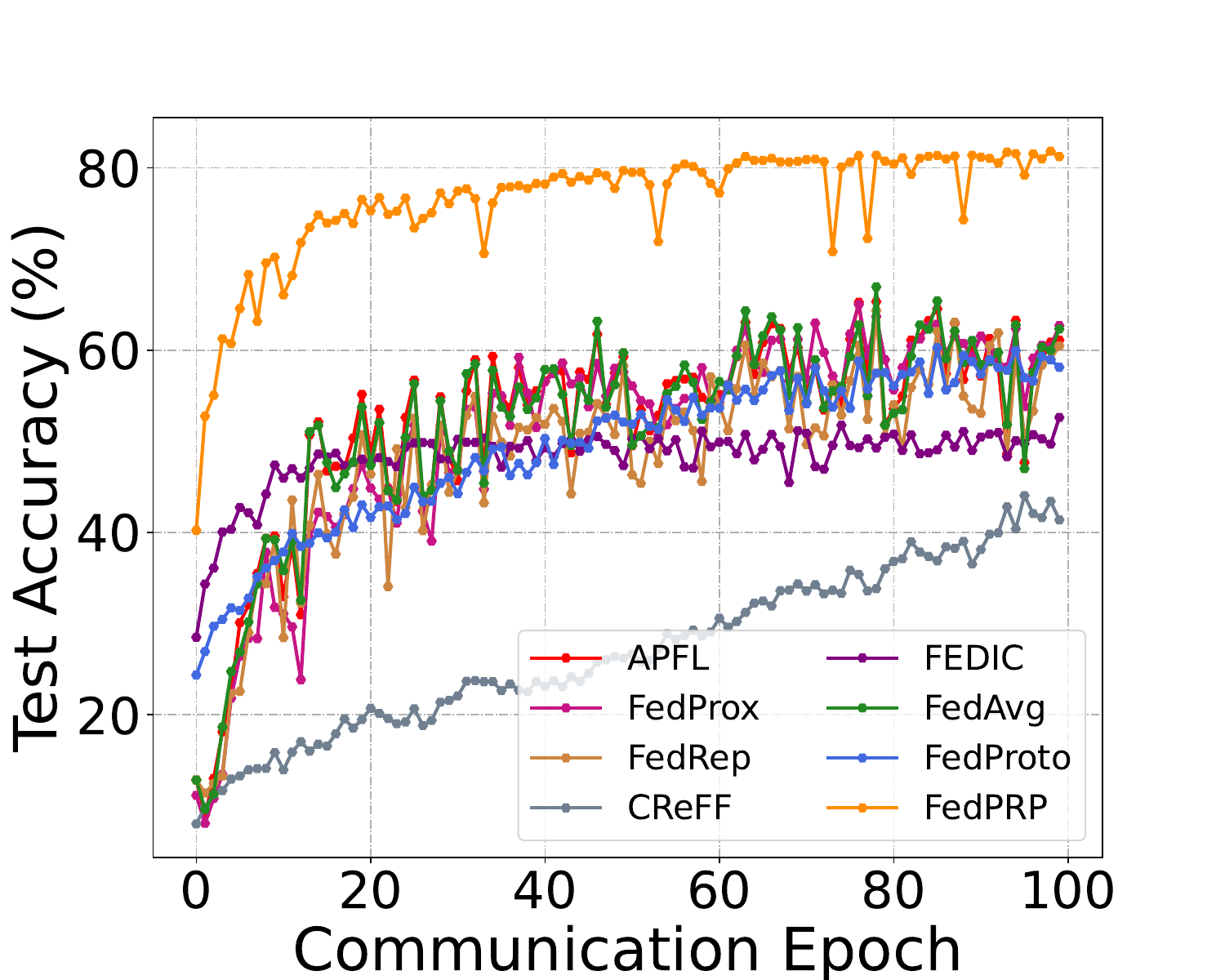}}
			\subfloat[$s$ = 5, CIFAR10]{\label{figure_cb}\includegraphics[height=0.2\textwidth]{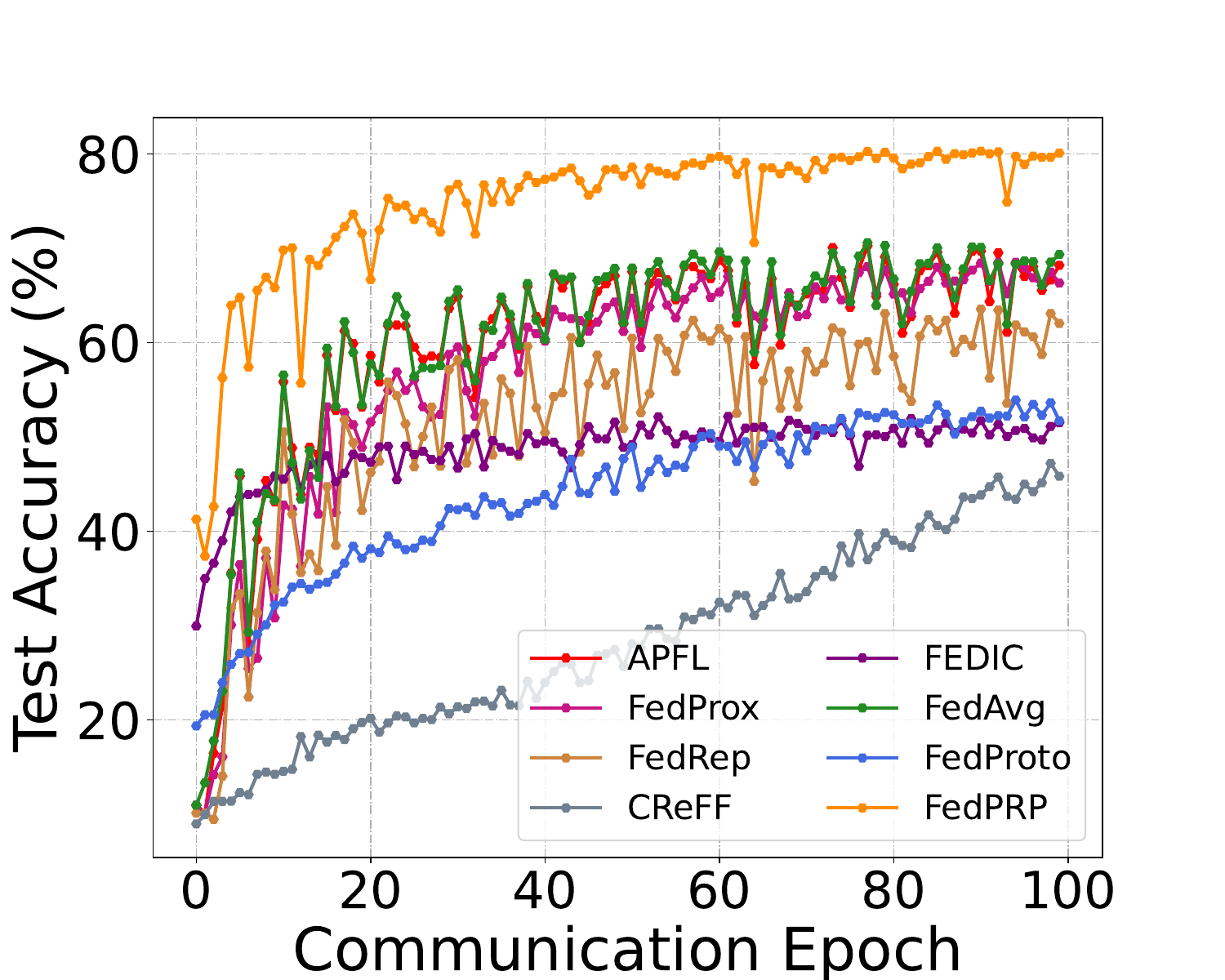}}
			\subfloat[$s$ = 20, CIFAR100]{\label{figure_cc}\includegraphics[height=0.2\textwidth]{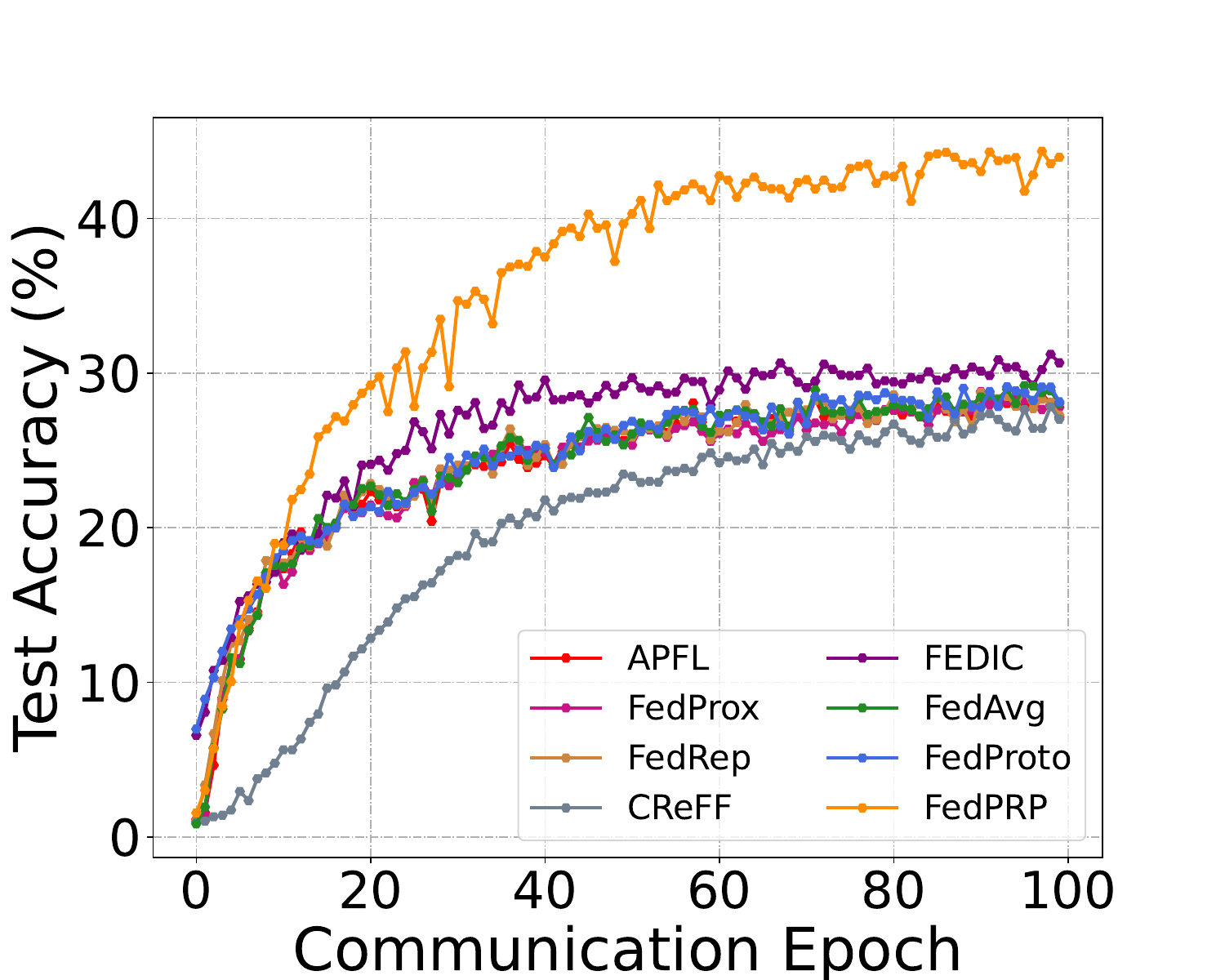}}
			\subfloat[$s$ = 30, CIFAR100]{\label{figure_cc}\includegraphics[height=0.2\textwidth]{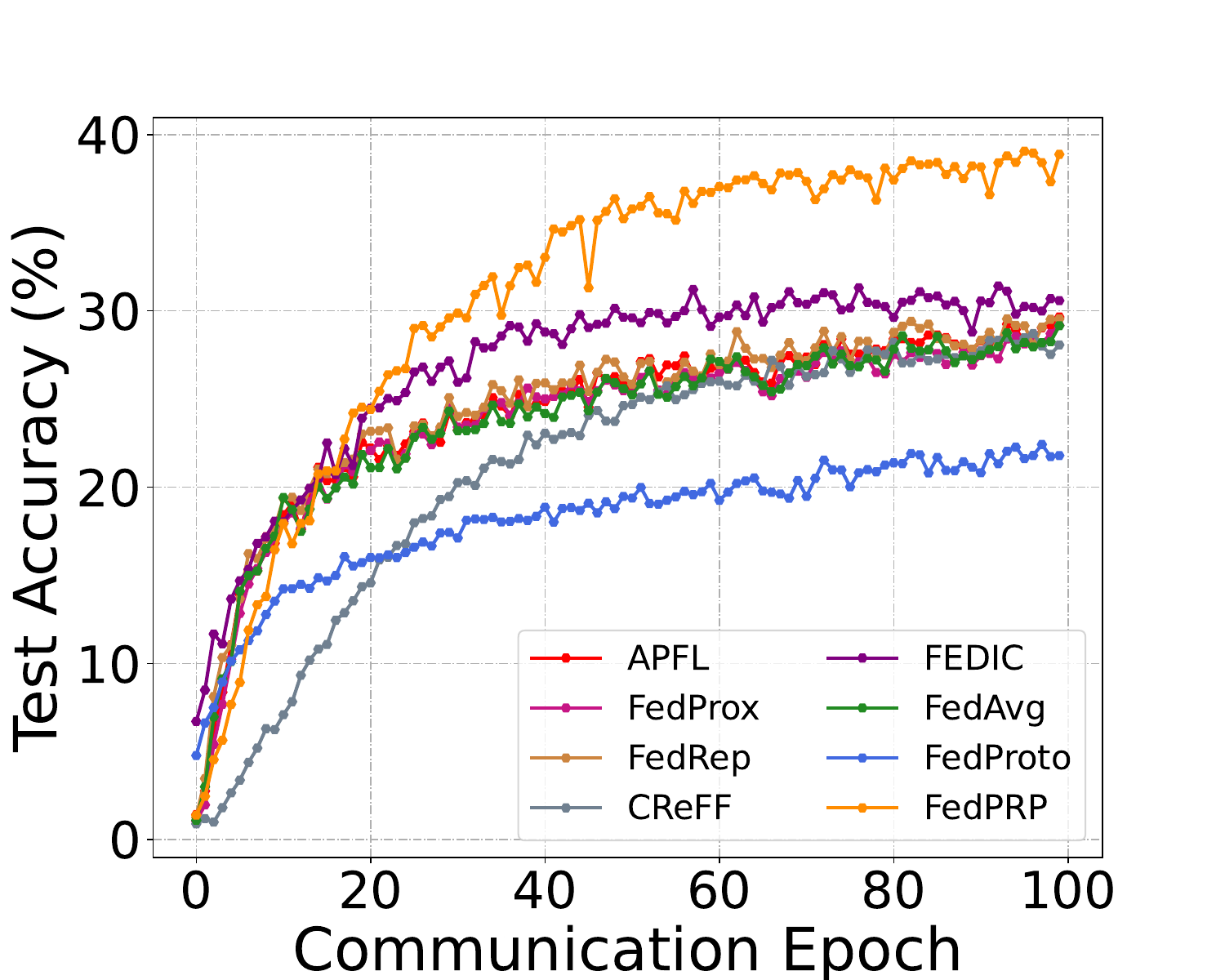}}
			\caption{The $\mathcal{A}^{glo}$ (\%) curves on CIFAR10 and CIFAR100 datasets with $\gamma$ = 0.5 under the \textit{Sharding} non-iid partition strategy.}
			\label{fig:globalmodel}}
	\end{figure*}
	
	\begin{table}[!htb]\footnotesize
		\centering
		\caption{The $\mathcal{A}^{sel}$ (\%) of different FL methods on CIFAR10 dataset under the \textit{DDA} non-iid partition strategy.}
		\resizebox{1\linewidth}{!}{
			\begin{tabular*}{9cm}{@{\extracolsep{\fill}}cccc|ccc}
				\toprule
				&\multicolumn{3}{c}{$\alpha$ = 0.4}&\multicolumn{3}{c}{$\alpha$ = 0.05}\\
				\midrule
				$\gamma$& 0.1 & 0.5 &1.0&0.1 & 0.5 &1.0 \\
				\midrule
				APFL~\cite{deng2020adaptive}     & 43.31 & 45.29 & 52.06 & 54.28 & 46.36 &   55.09    \\
				FedRep~\cite{collins2021exploiting}   & 42.44 & 50.09 & 50.36 & 57.14 & 49.64 & 55.25 \\
				FedProx~\cite{li2020federated}  & 42.92 & 55.64 & 61.79 & 53.68 & 51.79 &55.82\\
				FedProto~\cite{tan2022fedproto} & 37.78 & 37.84 & 38.22 & 54.07 & 50.22 & 55.36 \\
				FedAvg~\cite{mcmahan2017communication}   & 42.35 & 49.63 & 52.98 & 59.38 & 54.35 & 58.07 \\
				\midrule
				FEDIC~\cite{shang2022fedic}    &\underline{60.26} & \underline{60.07} & \underline{65.21} & \underline{66.02} & \underline{65.42} &\underline{ 69.58 }\\
				CReFF~\cite{2022Federated}    & 52.55 & 56.03 & 58.47 & 64.78 & 61.54 & 65.52 \\
				\midrule
				FedPRP  &\textbf{60.65}&\textbf{67.04}&\textbf{71.76}&\textbf{74.66}&\textbf{75.79}&\textbf{79.41}\\
				\bottomrule
		\end{tabular*}}
		\label{tab:local}
	\end{table}
	Under the \textit{DDA} non-iid partition strategy, FedPRP again shows superior performance in the \textit{All} metric and achieves substantial gains in the \textit{Medium} and \textit{Few} group classes. 
	Our method demonstrates superior performance across different class groups compared to addressing federated long-tailed learning with CReFF, particularly with a notable 2.35\% improvement observed in the \textit{Few} class. 
	
	\Cref{tab:splitcifar100} shows that the results on the CIFAR100 dataset exhibit a similar trend to that of CIFAR10. The proposed FedPRP obtains promising outcomes in both non-iid strategies. 
	In particular, under the \textit{Sharding} non-iid strategy, FedPRP exhibits a 5.69\% improvement over the FEDIC method on \textit{\textbf{All}}. It also achieves optimal performance on the \textit{Few} class, which is about 7.58\% to 10.40\% better than the personalized method, and also slightly improved compared to the CReFF and FEDIC methods.
	The results suggest that the FedPRP strategy of personalization and prototype rectification is significantly beneficial in skew heterogeneity settings.
	
	{\color{black}\subsubsection{Running Time} In Table~\ref{tab:time}, we report the running time per epoch of different methods on 10 participating clients, including training local models, validating test data according to class distributions, and interacting with the server. Personalized federated learning FedRep exhibits the shortest runtime due to its differentiation between personalized and shared parts of model training, emphasizing updates to a minority of personalized layers to reduce training time. Conversely, the proposed method incurs time due to the need for each client to update experience prototypes, in addition to server-side aggregation of prototypes. While the time spent on server-side prototype aggregation is comparable to model aggregation, the primary time consumption lies in the training of experience prototypes on each client, making it suitable for distributed scenarios with client-side parallelism.}

	\begin{table}[!htb]\footnotesize
		{\color{black}
			\centering
			\caption{Running time of per round for different FL methods under the \textit{DDA} non-iid partition strategy with $\gamma = 0.1$ and $\alpha$ = 0.4.}
			\resizebox{1\linewidth}{!}{
				\begin{tabular*}{9cm}{@{\extracolsep{\fill}}cccc}
					\toprule
					& \multicolumn{3}{c}{\textbf{computation and communication time (s)}}\\
					\cmidrule{2-4}
					Method & CIFAR10& CIFAR100 & Tiny-ImageNet\\
					\midrule
					APFL~\cite{deng2020adaptive}&370.85&386.76&59,482.37 \\
					FedRep~\cite{collins2021exploiting}&165.55 &169.81&27,683.77 \\
					FedProx~\cite{li2020federated}&194.26&197.73&32,338.43 \\
					FedProto~\cite{tan2022fedproto}&356.16&400.76&65,261.64 \\
					FedAvg~\cite{mcmahan2017communication}&191.12&192.51&33,654.78 \\
					\midrule
					FEDIC~\cite{shang2022fedic}&167.85&337.31&54,023.65\\
					CReFF~\cite{2022Federated}&277.11&458.47&71,866.20\\
					\midrule
					FedPRP &193.44&344.73&56,136.13\\
					\bottomrule
			\end{tabular*}}
			\label{tab:time}}
	\end{table}

	\subsubsection{Personalized Performance of Local Model}
	To validate the effectiveness of FedPRP in training personalized local models with limited sample sizes, we used test data corresponding to the class distribution for verification purposes. Table~\ref{tab:local} shows the personalized modeling capability of FedPRP, which consistently achieves the best results under varying degrees of skewed heterogeneity.
	
	Specifically, when the global data distribution is balanced with $\gamma$ = 1.0 and $\alpha$ = 0.4, FedPRP outperforms the suboptimal FEDIC method by an impressive margin of 6.55\%. Notably, as the heterogeneity increases (i.e., when $\alpha$ = 0.05), the overall performance exceeds that achieved with $\alpha$ = 0.4. This phenomenon can be attributed to the reduced number of classes within each client, resulting in a more concentrated set of learned classes, which facilitates better classification.
	
	\begin{figure}[!tb]
		\begin{center}
			\includegraphics[width=1\linewidth]{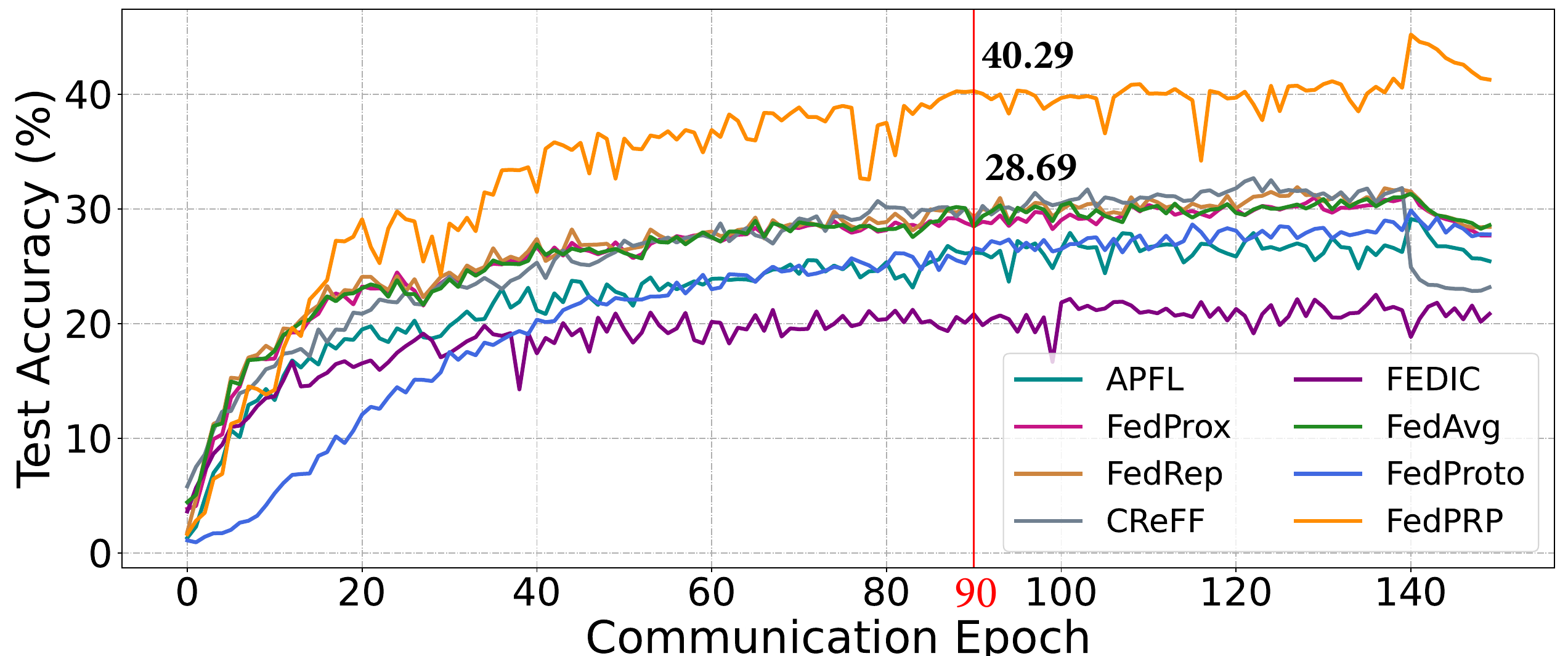}
		\end{center}
		\caption{Comparison of $\mathcal{A}^{loc}$ (\%) curves for all methods after adding 10 new clients, where $\gamma$ = 0.5 and $s$ = 20 on the CIFAR100 dataset.}
		\label{fig:ini}
	\end{figure}
	
	\subsubsection{Generalization Performance of the Global Model}
	To demonstrate the performance of the aggregated global model, we utilize the balanced test set for validation, as shown in \Cref{fig:globalmodel}. All algorithms show an increasing trend in accuracy as the number of communication epochs increases, indicating progressive learning. FedPRP consistently achieves superior performance across all \textit{sharding} levels on both datasets, underscoring the effectiveness of its prototype rectification and personalization strategy in SHFL.
	Especially on the CIFAR100 dataset with $s$ = 30, FedPRP exhibits a significant competitive advantage and even displays outstanding performance in the early stages of training. The observed oscillation phenomenon indicates significant differences in label distribution between the current and previous rounds of partial participants.
	
	To prove that the global training model trained by FedPRP generalizes, we use it to initialize ten new client models whose label distributions are all inconsistent when $epoch=90$.
	\Cref{fig:ini} illustrates the accuracy curve for predicting the balance test set using local models.
	It is obvious that the proposed method has a significant advantage from the beginning.
	The average $\mathcal{A}^{loc}$ of 10 new clients is 40.29\%, which is 11.60\% higher than the CReFF method.
	With the arrival of new participants, our proposed method is still in the best upward trend, which reflects that the initialized global model has certain generalizations.
	The decrease in accuracy from CIFAR10 to CIFAR100 and Tiny-ImageNet for all methods suggests increased difficulty due to more classes and potentially more complex data distributions. Despite this, FedPRP exhibits a relatively smaller drop in performance, highlighting its potential in complex federated learning scenarios.
	\begin{figure*}[!htb]
		\centering
		\subfloat[CIFAR10, $s$ = 4]{\label{figure_a}\includegraphics[height=0.2\textwidth]{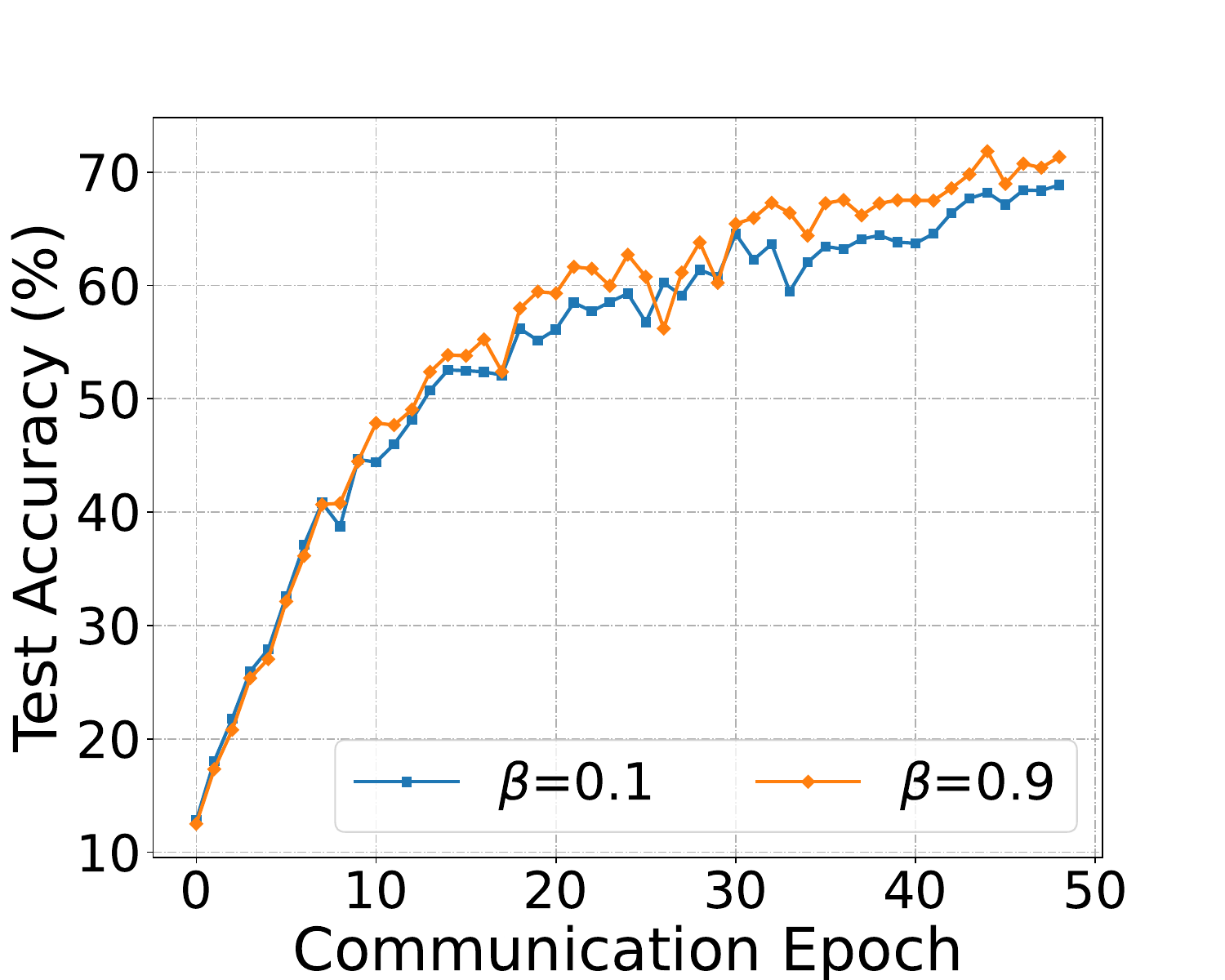}}
		\subfloat[CIFAR10, $s$ = 4]{\label{figure_b}\includegraphics[height=0.2\textwidth]{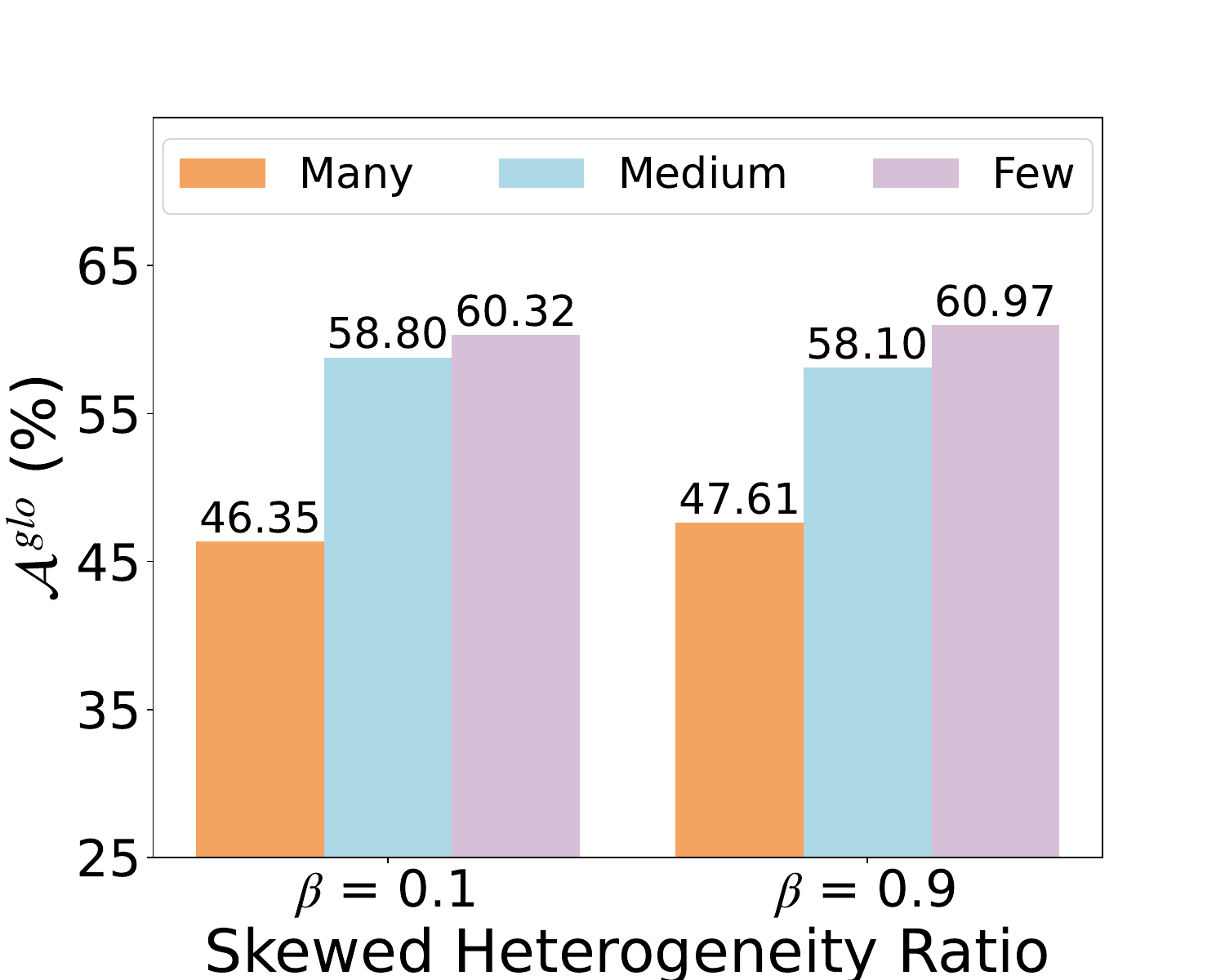}}
		\subfloat[CIFAR100, $s$ = 20]{\label{figure_c}\includegraphics[height=0.2\textwidth]{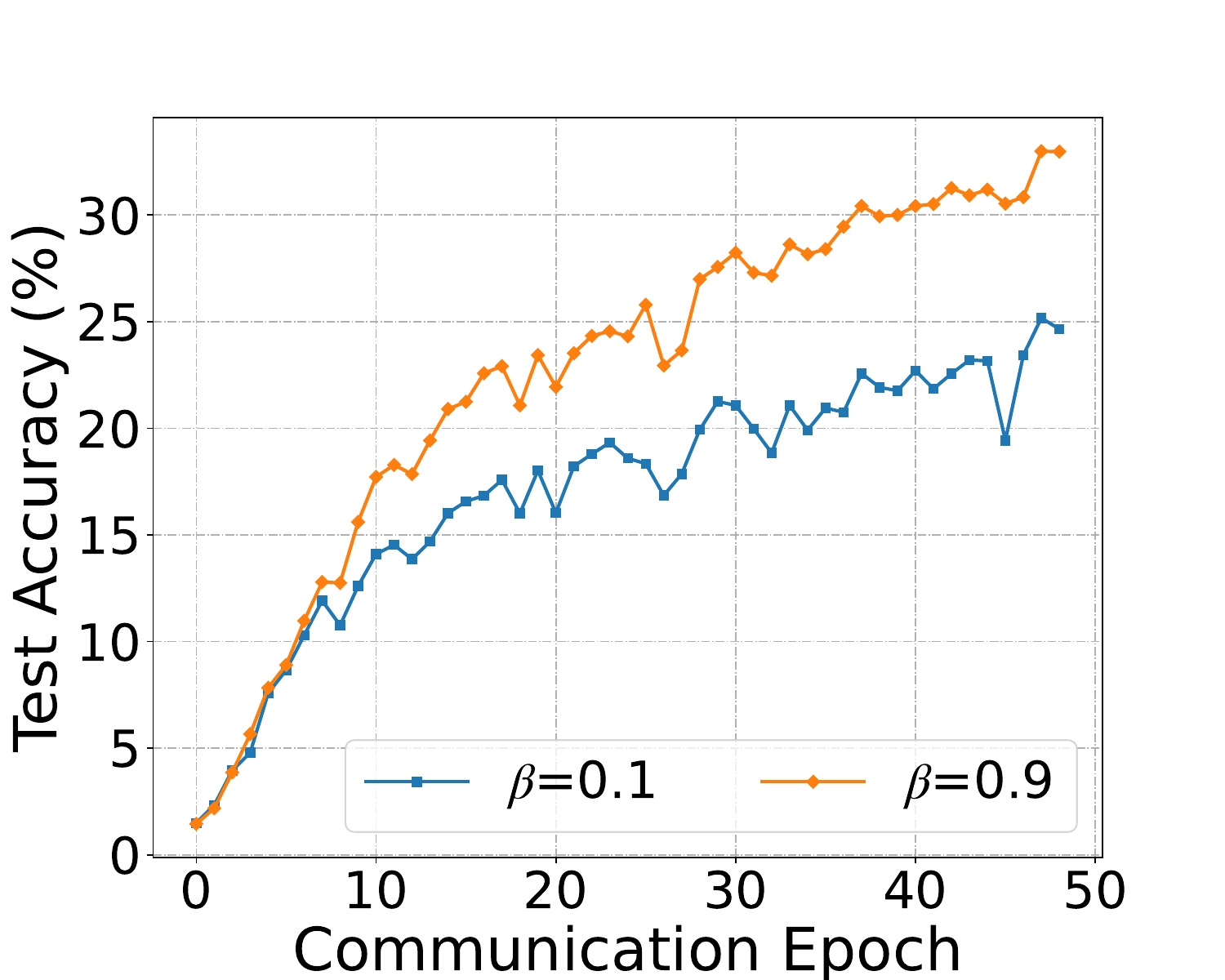}}
		\subfloat[CIFAR100, $s$ = 20]{\label{figure_d}\includegraphics[height=0.2\textwidth]{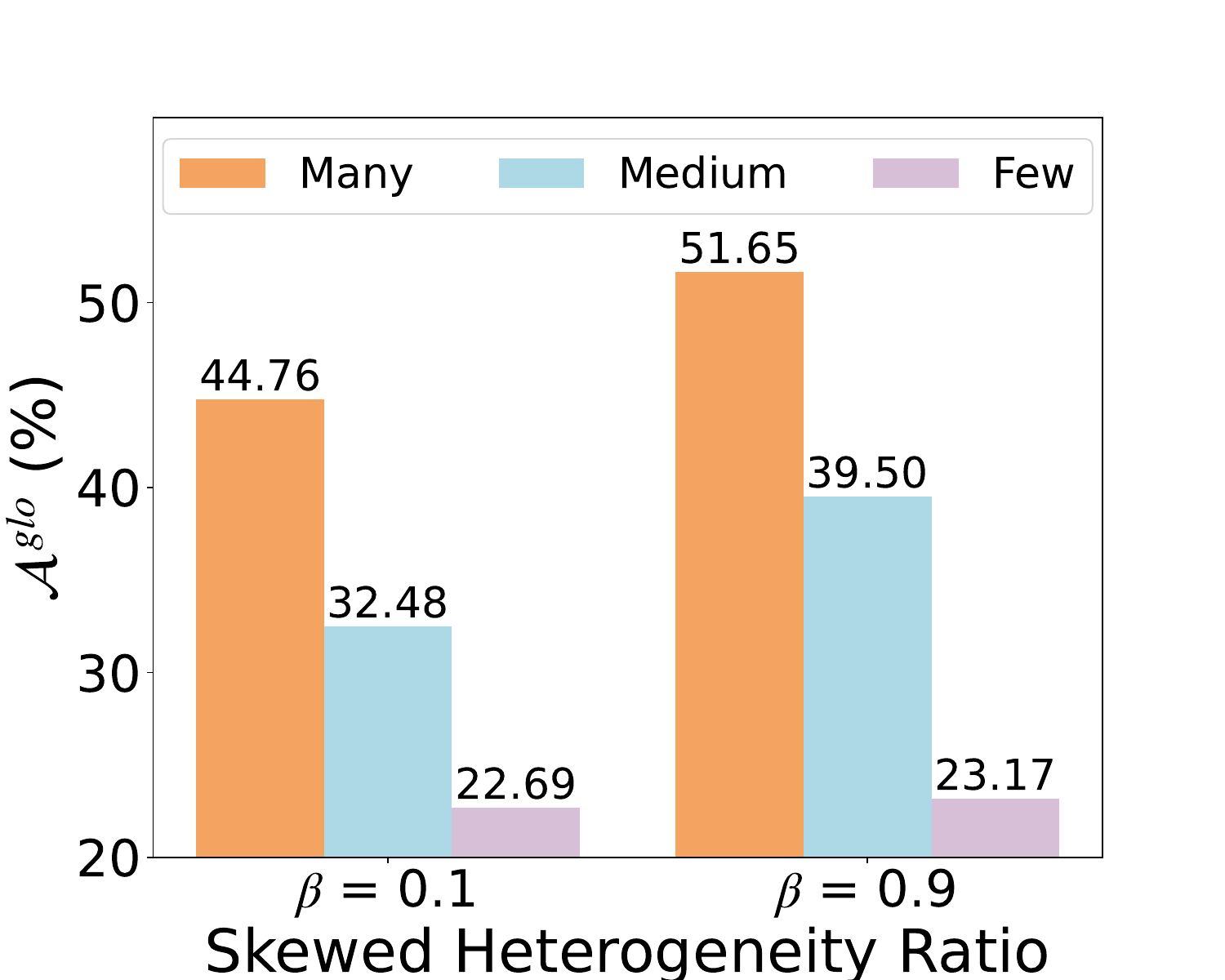}}
		\caption{Take different moving average parameters $\beta$ on the $\mathcal{A}^{glo}$ curves and $\mathcal{A}^{glo}$ values in three groups class with $\gamma$ = 0.1.}
		\label{fig:ablation}
	\end{figure*}
	
	Furthermore, \Cref{tab:rapid} presents a performance comparison of the different methods after the addition of 10 new clients and one round of training on their respective private data. Obviously, FedPRP shows consistently superior performance on all datasets. In particular, on CIFAR100, our method outperforms CReFF by 9.57\% on $\mathcal{A}^{loc}$.
	Meanwhile, the performance test results of the local and global models of FedPRP show relatively small differences, reflecting that the prototype representation contributes to the trade-off between the personalization of the local models and the generalization of the global model.
	
	\subsection{Ablation Studies}\label{section:ablation}
	\subsubsection{Analysis of Moving Average}
	To validate the effectiveness of using the moving average to update global prototypes on the server-side, we conducted experiments with various moving average values $\beta$, as shown in~\Cref{fig:ablation}. 
	\begin{table}[!tb]
		\centering
		\caption{The first round of test accuracy after ten new clients initialize the local model with $\gamma$ = 0.5 under the \textit{Sharding} non-iid partition strategy.}
		\resizebox{1\linewidth}{!}{
			\begin{tabular*}{9cm}{@{\extracolsep{\fill}}c c c |c c| c c }
				\toprule
				& \multicolumn{2}{c}{CIFAR10}& \multicolumn{2}{c}{CIFAR100}& \multicolumn{2}{c}{Tiny-ImageNet}\\
				
				& \multicolumn{2}{c}{$s$ =  5}& \multicolumn{2}{c}{$s$ =  20}& \multicolumn{2}{c}{$s$ =  20}\\
				\cmidrule{1-7}
				Metric & $\mathcal{A}^{loc}$  & $\mathcal{A}^{glo}$ & $\mathcal{A}^{loc}$ & $\mathcal{A}^{glo}$  & $\mathcal{A}^{loc}$ & $\mathcal{A}^{glo}$\\
				\midrule
				APFL~\cite{deng2020adaptive} &58.63&62.46&25.47&28.06&17.91&18.06\\
				FedRep~\cite{collins2021exploiting} &59.63&58.90&29.61&27.97&20.34&17.38 \\
				FedProx~\cite{li2020federated} & 66.36&65.36&29.96&28.04&22.45&19.71\\
				FedProto~\cite{tan2022fedproto} &45.42&45.84&28.54&\underline{28.69}&27.57&21.45\\
				FedAvg~\cite{mcmahan2017communication} &64.43&63.68&29.98&28.48&22.06&19.52\\
				\midrule
				FEDIC~\cite{shang2022fedic} &60.76&60.07&31.47&25.78&\underline{38.57}&\underline{37.01}\\
				CReFF~\cite{2022Federated} &\underline{61.66}&\underline{61.88}&\underline{33.52}&26.66&21.99&19.80 \\
				\midrule
				FedPRP & \textbf{77.89}&\textbf{78.43}&\textbf{41.09}&\textbf{41.17}&\textbf{40.47}&\textbf{39.96}\\
				\bottomrule
		\end{tabular*}}
		\label{tab:rapid}
	\end{table}
	
	Figures \ref{figure_a} and \ref{figure_c} depict curves for different settings of the parameter $\beta$, with $\beta = 0.9$ demonstrating a marked prominence over $\beta$ = 0.1. This highlights that smoothly integrating new information while retaining useful knowledge is more conducive to approaching the class centers, thereby improving classification performance. In Figures \ref{figure_b} and \ref{figure_d}, the validation results over three group classes also show a pronounced effectiveness when $\beta = 0.9$. This is particularly evident in the CIFAR100 dataset, where improvements are observed in all three class groups, most notably a 6.89\% increase in the \textit{Many} class. These findings underscore the effectiveness of using moving averages to smoothly update global prototypes.
	
	{\color{black}\subsubsection{Analysis of Components} To demonstrate the contributions of the two components $\mathcal{L}_{ID}$ and $\mathcal{L}_{\mathrm{IC}}$ in FedPRP, a series of ablation experiments were performed as listed in~\Cref{tab:ab}. Compared to scenarios where one component was ablated, the complete FedPRP model maintained a significant advantage in results, underscoring the importance of each component in the framework. With $s$ = 20 and $\gamma$ = 0.5 on the CIFAR100 dataset, FedPRP outperformed ``w/o $\mathcal{L}_{ID}$" by 3.54\% on $\mathcal{A}^{loc}$. 
		The performance of ``w/o $\mathcal{L}_{ID}$" surpasses that of ``w/o $\mathcal{L}_{ID}$", highlighting the effectiveness of the inter-class discrimination loss. This underscores its ability to draw samples of minority classes closer to their respective centroids while distancing them from other classes, thereby improving classification performance. This illustrates the effectiveness of prototype-based learning training, which can enhance the training of shared representations to obtain high-quality classification representations. Classification using relatively balanced prototypes can effectively mitigate performance degradation in skewed class distribution scenarios.}

	\begin{table}[t]
		\label{ab}
		\centering{\color{black}
			\caption{Ablation studies for the proposed FedPRP with $\gamma$ = 0.5 on the CIFAR10 and CIFAR100. }
			\label{tab:ab}
			\resizebox{1\linewidth}{!}{
				\begin{tabular*}{9cm}{@{\extracolsep{\fill}}l|cc|cc|cc}
					\toprule
					\multirow{2}{*}{\textbf{Settings}} &	\multirow{2}{*}{$\mathcal{L}_{\mathrm{ID}}$} &	\multirow{2}{*}{$\mathcal{L}_{\mathrm{IC}}$} & \multicolumn{2}{c|}{$s = 5$} & \multicolumn{2}{c}{$s = 20$}  \\
					
					& &&$\mathcal{A}^{loc}$ & $\mathcal{A}^{glo}$ & $\mathcal{A}^{loc}$ & $\mathcal{A}^{glo}$ \\
					\midrule
					Baseline & \XSolidBrush & \XSolidBrush &60.95&60.83&22.67 &22.58\\
					FedPRP-w/o $\mathcal{L}_{\mathrm{ID}}$ & \XSolidBrush & \Checkmark &79.21&78.23&39.98&40.93 \\
					FedPRP-w/o $\mathcal{L}_{\mathrm{IC}}$ & \Checkmark & \XSolidBrush & 73.54&72.83&32.47 &29.87\\
					FedPRP (Ours) & \Checkmark & \Checkmark &\textbf{79.73}&\textbf{79.20}&\textbf{43.52}&\textbf{42.56} \\
					\bottomrule
		\end{tabular*}}}
		
	\end{table}
	
	\section{Conclusion}
	
	In this paper, we briefly review the data heterogeneity of federated learning and reformulate a challenging setting, i.e., Skewed Heterogeneous Federated Learning. To address this problem, we propose a novel learning framework named Federated Prototype Rectification with Personalization (FedPRP). 
	Empirical evaluations reveal that in this setting, FedPRP ensures the effectiveness of dominant classes while augmenting the classification performance of minority classes. Further analysis shows that it could adeptly balance the personalized effectiveness of local models with the generalization ability of the global model. The ablation study also highlights the indispensability and complementarity of the components of the proposed model.
	In the future, we will delve into the FedPRP framework in real-time dynamic federated learning environments, where data is continuously generated and models need to quickly adapt to the changing data distribution.

	\bibliographystyle{IEEEtran}
	\bibliography{ref}
	
\end{document}